\documentclass[letterpaper,journal]{IEEEtran}
\usepackage{amsmath,amsfonts}
\usepackage{algorithmic}
\usepackage{algorithm}
\usepackage{array}
\usepackage[caption=false,font=normalsize,labelfont=sf,textfont=sf]{subfig}
\usepackage{textcomp}
\usepackage{stfloats}
\usepackage{url}
\usepackage{verbatim}
\usepackage{graphicx}
\usepackage{cite}
\def\ie{{\emph{i.e.}}}
\def\etc{{\emph{etc.}}}
\def\etal{{\emph{et al.}}}
\usepackage{graphicx}
\usepackage{amsmath}
\usepackage{amssymb}
\usepackage{booktabs}
\usepackage{pifont} % colors
\usepackage{amsmath}
\usepackage{amssymb}
\usepackage{graphicx}
\usepackage{array}
\usepackage{comment}
\usepackage{epsfig}
\usepackage{times}
\usepackage{comment}
\usepackage{url}
\usepackage{subfig}
\usepackage{times}
\usepackage{algorithm,algorithmic,xspace}
\usepackage{verbatim}
\usepackage{multirow}
\usepackage{colortbl}
\usepackage{bbding}
\usepackage{wasysym}
\usepackage{amssymb}
\usepackage[table]{xcolor}
\usepackage{breqn}
\usepackage{graphicx,animate}
\usepackage[normalem]{ulem}
\usepackage{tikz}
\usepackage{flushend}
\newcommand*{\circled}[1]{\lower.7ex\hbox{\tikz\draw (0pt, 0pt)%
    circle (.4em) node {\makebox[1em][c]{\small #1}};}}

\usepackage{bigstrut}
\newcommand{\Tref}[1]{Table~\ref{#1}}
\newcommand{\Eref}[1]{Equation~(\ref{#1})}
\newcommand{\Fref}[1]{Fig.~\ref{#1}}
\newcommand{\Sref}[1]{Section~\ref{#1}}

\def\etal{\emph{ et al.}}
\def\ie{\emph{i.e.}}

\def\ie{{\emph{i.e.}}}
\def\etc{{\emph{etc.}}}
\def\etal{{\emph{et al.}}}

\def\Pi{{\mathbf{M}_{i}}}

\definecolor{Auqamarin}{gray}{0.9}
\definecolor{LightCyan}{rgb}{0.88,1,1}
\definecolor{myGreen}{rgb}{0, .9, .6}

\definecolor{americanrose}{rgb}{1.0, 0.01, 0.24}

\begin{document}

% \title{MessageNeRF: Embedding Messages into Neural Radiance Fields for Copyright Protection}

\title{The NeRF Signature: Codebook-Aided Watermarking for Neural Radiance Fields}

\author{
        Ziyuan Luo, \textit{Student Member}, \textit{IEEE},
        Anderson Rocha, \textit{Fellow}, \textit{IEEE},
        Boxin Shi, \textit{Senior Member}, \textit{IEEE},\\
        Qing Guo, \textit{Senior Member}, \textit{IEEE},
        Haoliang Li, \textit{Member}, \textit{IEEE},        
        Renjie Wan*, \textit{Member}, \textit{IEEE}
\thanks{This work was done at Renjie Group at Hong Kong Baptist University.}
\thanks{Ziyuan Luo and Renjie Wan are with the Department of Computer Science, Hong Kong Baptist University, Hong Kong SAR, China (e-mail: ziyuanluo@life.hkbu.edu.hk, renjiewan@hkbu.edu.hk).}% <-this % stops a space
\thanks{Anderson Rocha is with the Institute of Computing, University of Campinas, Brazil.}%
\thanks{Boxin Shi is with the State Key Laboratory of Multimedia Information Processing and National Engineering Research Center of Visual Technology, School of Computer Science, Peking University, Beijing, 100871, China.}%
\thanks{Qing Guo is with A*STAR, Singapore.}%
\thanks{Haoliang Li is with the Department of Electrical Engineering, City University of Hong Kong, Hong Kong.}%
\thanks{* Corresponding author: Renjie Wan.}%
}

% The paper headers
\markboth{IEEE Transactions on Pattern Analysis and Machine Intelligence}%
{Shell \MakeLowercase{\textit{et al.}}: A Sample Article Using IEEEtran.cls for IEEE Journals}

% \IEEEpubid{0000--0000/00\$00.00~\copyright~2021 IEEE}
% Remember, if you use this you must call \IEEEpubidadjcol in the second
% column for its text to clear the IEEEpubid mark.

\maketitle

\begin{abstract}

Neural Radiance Fields (NeRF) have been gaining attention as a significant form of 3D content representation. With the proliferation of NeRF-based creations, the need for copyright protection has emerged as a critical issue. Although some approaches have been proposed to embed digital watermarks into NeRF, they often neglect essential model-level considerations and incur substantial time overheads, resulting in reduced imperceptibility and robustness, along with user inconvenience. In this paper, we extend the previous criteria for image watermarking to the model level and propose NeRF Signature, a novel watermarking method for NeRF. We employ a Codebook-aided Signature Embedding (CSE) that does not alter the model structure, thereby maintaining imperceptibility and enhancing robustness at the model level. Furthermore, after optimization, any desired signatures can be embedded through the CSE, and no fine-tuning is required when NeRF owners want to use new binary signatures. Then, we introduce a joint pose-patch encryption watermarking strategy to hide signatures into patches rendered from a specific viewpoint for higher robustness. In addition, we explore a Complexity-Aware Key Selection (CAKS) scheme to embed signatures in high visual complexity patches to enhance imperceptibility. The experimental results demonstrate that our method outperforms other baseline methods in terms of imperceptibility and robustness. The source code is available at: \url{https://github.com/luo-ziyuan/NeRF_Signature}.

\end{abstract}

\begin{IEEEkeywords}
Neural radiance fields, 3D reconstruction, digital watermarking.
\end{IEEEkeywords}

\section{Introduction}
\label{sec:intro}
\IEEEPARstart{N}{}eural Radiance Fields (NeRF) have become an important form of 3D digital assets. Many NeRFs have been created and publicly shared. Protecting the copyright of these created models is crucial to prevent unauthorized misuse and illegal transactions.

Generally, copyright protection for digital assets can be achieved through digital watermarking~\cite {zhu2018hidden, zhang2021deep, zhang2020udh}. Previous watermarking methods are mainly designed for 2D images and comply with two common standards~\cite{zhang2020udh, jing2021hinet}: 1) \textbf{imperceptibility}, which ensures that embedded signatures do not cause significant visual degradation; and 2) \textbf{robustness}, which requires the reliable extraction of signatures against various distortions.

\begin{figure}
  \centering
  \includegraphics[width=\linewidth]{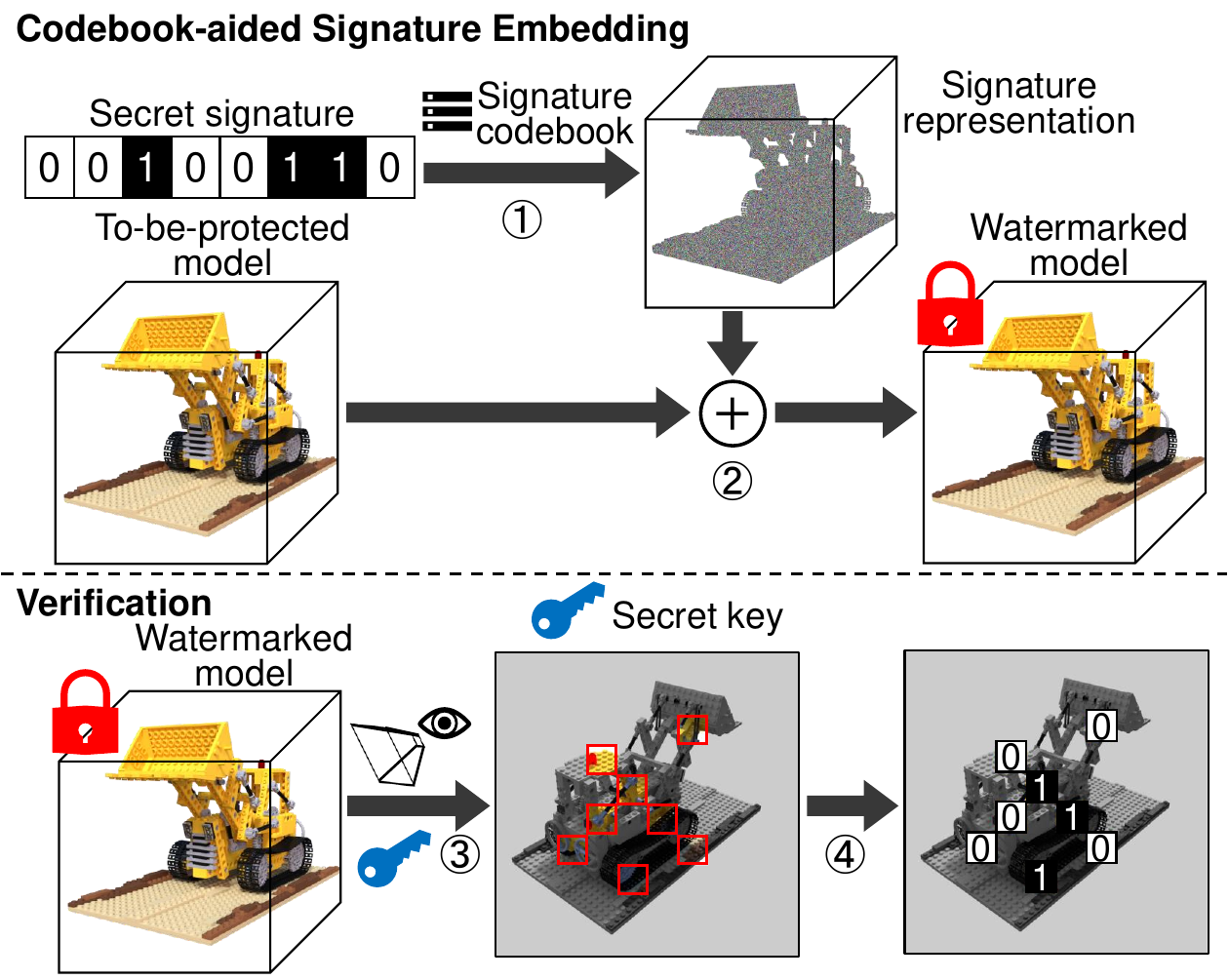}
  \caption{The embedding and verification of our NeRF Signature with an optimizable signature codebook. The NeRF owner first obtains a signature representation through the signature codebook with a selected secret signature (\ding{192}). Then, the watermarked model is created through element-wise addition between the signature representation and the original parameters while maintaining the model structure (\ding{193}). After sharing the watermarked model, the NeRF owner can render specific patches from a specific viewpoint using the secret key (\ding{194}). Finally, the signature can be extracted from these patches by an extractor (\ding{195}).}  

  \label{fig:fig1}
\end{figure}

Our previously proposed CopyRNeRF~\cite{luo2023copyrnerf} is the first attempt for NeRF watermarking. CopyRNeRF~\cite{luo2023copyrnerf} embeds binary signatures into the radiance fields by constructing a watermarked color representation. Another strategy is to embed a fixed signature by fine-tuning the original NeRF via an external module~\cite{jang2024waterf}, referred to as the fine-tuning pipeline.  Due to the difficulty of extracting embedded signatures from radiance fields, both pipelines ensure that these signatures are transmitted into every rendered image during volume rendering. Subsequently, an external signature extractor is used to extract binary signatures from the rendered images for copyright verification.

Despite the progress made by the two pipelines, they both reveal significant limitations. First, the two pipelines rely on external modules to complete signature embedding and extraction. If malicious users access those external modules, they can easily obtain information related to the embedded signatures, which can then be used to remove the copyright information. For example, CopyRNeRF~\cite{luo2023copyrnerf} encodes the binary signatures via separate modules, which can be easily detected and removed by malicious users. Signature extractors in the two pipelines may all be utilized by malicious users to detect and diminish the signatures. Second, the external modules and the core NeRF model cannot achieve compatibility, making the signature embedding difficult. To address this issue, CopyRNeRF~\cite{luo2023copyrnerf} employs several feature fusion strategies, albeit with high computational costs. The fine-tuning pipeline requires modifying all parameters of NeRF, which can easily degrade the model's quality and compromise the imperceptibility of watermarks.

Furthermore, fast training and rendering speed are important trends in the evolution of NeRF~\cite{fridovich2022plenoxels, muller2022instant, barron2023zip}. For the fine-tuning pipeline, when the NeRF owner needs to update the embedded signatures, the NeRFs should undergo another round of fine-tuning.  This can result in inconvenience for users and incur extra costs in terms of fine-tuning time. Though CopyRNeRF~\cite{luo2023copyrnerf} can address this issue by including all the potential signatures in the training phase, it stores information containing all signatures in the external MLP-based modules. This vast amount of information significantly increases the burden on the training process for signature embedding. The additional burdens may significantly undermine their usability and practicability.

One reason for the above limitations is that these methods primarily focus on the two standards at the image level. The ignorance of requirements at the core model level leads to poor performance in model compatibility, computational efficiency, and signature flexibility. Therefore, in addition to guaranteeing the two standards on rendered images, the modules used for signature embedding and extraction should not alter the core structures of NeRF and should remain resilient to malicious attacks. Furthermore, efficiency and convenience should be considered in the envisioned framework for practical deployment and real-world applications.

To achieve the above goals, we have the following considerations. First, considering the needs of practical application, we put convenience and efficiency of the envisioned framework foremost. The training process should be quick and efficient, and the NeRF owners should be able to flexibly update signatures, without requiring further fine-tuning. Second, the use of external modules during signature embedding and extraction should be minimized. Minimizing the number of external modules diminishes the potential for malicious attacks and ensures minor alterations to core NeRF models. Third, if some external modules are unavoidable, we propose to apply those unavoidable modules in a more encrypted manner to achieve higher model-level robustness. Then, even when malicious users access those unavoidable modules, it is still difficult to directly obtain the copyright signatures.

Our previous work, CopyRNeRF~\cite{luo2023copyrnerf}, allows model owners to update signatures without requiring another round of fine-tuning. If we can mitigate the challenges such as model-level threats and computational burdens brought by external modules, such convenience can better benefit the users.  To achieve this goal, we propose the NeRF Signature. Our \textbf{first} design is to employ a Codebook-aided Signature Embedding (CSE). Unlike CopyRNeRF~\cite{luo2023copyrnerf}, which relies on separate modules incompatible with core NeRF models, our CSE can add signature representations from an optimizable signature codebook directly with a portion of the NeRF model. The direct integration with the NeRF parameters fundamentally reduces potential vulnerabilities where malicious users could detect and remove external watermarking modules. Compared to the fine-tuning pipeline, only a portion of the NeRF model is utilized for embedding, minimizing interruptions to the information stored in NeRFs. \textbf{Second}, considering a binary signature with $N_{\text{b}}$ bit, such an optimizable signature codebook can represent all $2^{N_{\text{b}}}$ potential signatures for embedding, by accumulating the representations of each individual bit. When new signatures are required, NeRF owners can acquire signature representations using this codebook and directly add them to the NeRF parameters for embedding without the need for fine-tuning. This efficient representation significantly reduces the computational burden compared to previous methods that need to store signature information in external modules.
 \textbf{Third}, as the signature extractor is unavoidable for signature retrieval within the common digital watermarking framework~\cite{zhu2018hidden, ying2022rwn, zhou2022robust}, we propose a joint pose-patch encryption watermarking strategy to ensure that the signature can only be extracted from specific patches within a specific view, using a pose key and a patch key. Therefore, even if the extractor is accessed by malicious users, the secret signature can not be obtained without knowledge of the actual pose key and patch key. Lastly, to further enhance the imperceptibility of the watermarking, we explore a Complexity-Aware Key Selection (CAKS) scheme to select patches with high visual complexity as the embedded areas.

As outlined in~\Fref{fig:fig1}, once the codebook has been optimized, in the embedding stage, the NeRF owner can embed any desired binary signatures with a specific length into the model. In the verification stage, the NeRF owner can extract the hidden signature using the secret key. Throughout this entire procedure, the structure of the watermarked model remains identical to that of the to-be-protected model, ensuring it remains indistinguishable at the model level.

The \textbf{key contributions} of this paper are:
\begin{enumerate}
\item a novel watermarking method for NeRF with a Codebook-aided Signature Embedding (CSE). This embedding allows NeRF owners to embed desired signatures without altering the structure of the NeRF, thereby maintaining convenience while enhancing imperceptibility and robustness at the model level.

\item a joint pose-patch encryption watermarking strategy, which can embed signatures into patches rendered from a specific viewpoint to ensure the robustness of NeRF watermarking.

\item a Complexity-Aware Key Selection (CAKS) scheme, which can ensure the information is embedded into patches with high visual complexity to further strengthen imperceptibility.
\end{enumerate}

% \noindent{}{Comparison with CopyRNeRF}

Compared with CopyRNeRF~\cite{luo2023copyrnerf}, our NeRF Signature differs in the following key aspects. First, instead of using separate modules that are incompatible with core NeRF models, our method employs CSE to directly integrate signature representations into the NeRF model without structural modifications. Second, while CopyRNeRF~\cite{luo2023copyrnerf} stores complete signature information in external MLP-based modules, our method utilizes an optimizable signature codebook that efficiently represents signatures by accumulating bit-wise representations, significantly reducing storage and computational costs. Third, our method introduces additional security measures including the joint pose-patch encryption watermarking strategy and CAKS scheme. These features can prevent unauthorized signature extraction and ensure effective watermarking.

\section{Related Work}
\subsection{Neural Radiance Fields (NeRF)}
NeRF~\cite{nerf2020} is a revolutionary method that has emerged to provide high-quality scene representation by fitting a neural radiance field to a set of RGB images with corresponding poses. Vanilla NeRF involves querying a deep MLP model millions of times~\cite{reiser2021kilonerf}, leading to slow training and rendering speeds. Some research efforts have tried to speed up this process by using more efficient sampling schemes~\cite{piala2021terminerf, 10328666, verbin2024ref}, while some have attempted to apply improved data structures to represent the objects or scenes~\cite{10478788, muller2022instant, sun2022direct, fridovich2022plenoxels, chen2022tensorf}. Besides, to improve the NeRF training on low-quality images, enhancements have been made to handle degradation, such as blurring~\cite{ma2022deblur, wang2023bad, qi2023e2nerf, zhou2023nerflix}, lowlight~\cite{wang2023lighting}, and reflection~\cite{guo2022nerfren, zhu2022neural}. NeRF has been applied to a broader range of scenarios, including indoor scene reconstruction~\cite{10496207, wei2023depth, chen2023structnerf, yang2023nerfvs}, human body modeling~\cite{gao2022mps, peng2024animatable}, and 3D segmentation~\cite{cen2023segment, cen2023segment_pami, liu2024sanerf}. Recently, 3D Gaussian Splatting~\cite{kerbl3Dgaussians} has made significant progress in 3D scene representation, demonstrating its effectiveness in various domains including object reconstruction~\cite{charatan2024pixelsplat, chen2024liftimage3d}, medical applications~\cite{wang2024endogslam, li2024endosparse}, and avatar creation~\cite{qian20243dgs, zhao2024chase, zhao2024sg}. 
As NeRF-based 3D assets gain popularity among creators, protecting the copyright of these assets has become crucial.

\subsection{Digital watermarking for 2D images} 
2D digital watermarking is used for image verification, authenticity, and traceability. Initial 2D watermarking methods hide data in the least significant parts of image pixels \cite{413536}. Alternatively, some techniques embed data in transformed domains \cite{lai2010digital, kang2010efficient, wang2022dtcwt}. Recently, deep learning techniques have shown significant advancements in information hiding in images~\cite{zhu2018hidden, baluja2019hiding, jing2021hinet, ying2022rwn, zeng2023watermarks}. HiDDeN~\cite{zhu2018hidden} is one of the first deep image watermarking methods that employ deep encoder-decoder networks to achieve superior performance compared to traditional approaches. UDH~\cite{zhang2020udh} proposes a universal deep hiding architecture to achieve cover-agnostic watermark embedding. From then on, many methods have focused on more robust watermark embedding and extraction under distortion conditions, such as JPEG compression~\cite{jia2021mbrs}, screen recapture~\cite{fang2018screen, fang2022pimog, fang2023denol, liu2023wrap}, and combinations of several distortions~\cite{luo2020distortion}. Besides the encoder-decoder paradigm, some invertible networks have also been used for digital watermarking~\cite{guan2022deepmih, fang2023flow}. Recently, methods have been proposed to watermark the generative content~\cite{fernandez2023stable}. However, those 2D digital watermarking methods cannot be directly applied to protect the copyright of 3D models~\cite{luo2023copyrnerf}.

\subsection{Digital watermarking for 3D models}
Early 3D watermarking approaches~\cite{ohbuchi2002frequency, son2017perceptual, al2019graph} rely on Fourier or wavelet analysis on triangular or polygonal meshes to encode messages into model frequencies. However, these techniques take much time to work as the number of points in the model increases. Later approaches~\cite{zhou2018distortion, jiang2017reversible, tsai2022integrating, hou2023separable} suggest putting watermarks into the least significant bits and the most significant bits of vertex coordinates. Recently, some studies~\cite{wang2022neural, wang2022deep, zhu2024rethinking} have explored the feasibility of deep neural networks for 3D watermarking. Yoo \etal~\cite{Yoo_2022_CVPR} propose to embed messages in 3D meshes and extract them from 2D renderings. Although neural networks are commonly present in NeRF, watermarking methods designed for neural networks~\cite{zhang2024robust, tan2023deep, zeng2023watermarks, wu2020watermarking} cannot be directly applied to NeRF watermarking.

CopyRNeRF~\cite{luo2023copyrnerf} is the first method to watermark a NeRF. However, its long training time hinders its practicality in real-world scenarios. StegaNeRF~\cite{li2023steganerf} proposes to embed steganographic information within NeRF. In this method, the detector holds the most hidden information, which can make the information vulnerable to leaks, reducing its overall robustness. WateRF~\cite{jang2024waterf} leverages a fine-tuning technique for watermarking NeRFs, but it is limited to embedding just one signature following each fine-tuning process.

\section{Preliminaries}
\label{sec:preliminary}
We briefly introduce NeRF~\cite{nerf2020} and CopyRNeRF~\cite{luo2023copyrnerf} in this section. NeRF~\cite{nerf2020} uses MLPs to map the 3D location $\mathbf{x} \in \mathbb{R}^3$ and viewing direction $\mathbf{d} \in \mathbb{R}^2$ to a density value $\sigma \in \mathbb{R}^+$ and a color value $\mathbf{c} \in \mathbb{R}^3$:
\begin{equation}
\sigma(\mathbf{x})=f_{\sigma}\left(\gamma_{\mathbf{x}}(\mathbf{x}); \theta_{\sigma}\right),
\label{eq:sigma}
\end{equation}
\begin{equation}
\mathbf{c}(\mathbf{x}, \mathbf{d})=f_{\text{c}}\left(\gamma_{\mathbf{x}}(\mathbf{x}), \gamma_{\mathbf{d}}(\mathbf{d}); \theta_{\text{c}}\right),
\label{eq:nerffirst}
\end{equation}
where $\theta_{\sigma}$ and $ \theta_{\mathbf{c}}$ are the parameters of MLPs for representing density and color, respectively. $\gamma_{\mathbf{x}}$ and $\gamma_{\mathbf{d}}$ are the fixed positional encoding functions for location and viewing direction, respectively. The volumetric rendering equation is used to obtain the color in 2D images:
\begin{equation}
\mathbf{C}(\mathbf{r})=\int_{t_n}^{t_f} T(t) \sigma(\mathbf{r}(t)) \mathbf{c}(\mathbf{r}(t), \mathbf{d}) d t,
\label{eq:rendering}
\end{equation}
\begin{equation}
T(t)=\exp (-\int_{t_n}^t \sigma(r(s)) d s),
\label{eq:rendering2}
\end{equation}
where $\mathbf{r}(t)$ is a ray from a camera viewpoint, and $\mathbf{C}(\mathbf{r})$ is the expected color of the ray $r(t)$. The near and far integral bounds are denoted as $t_n$ and $t_f$, respectively. In practice, a numerical quadrature is used to approximate integral in~\Eref{eq:rendering}.

Given a to-be-watermarked NeRF with optimized $\theta_{\sigma}$ and $ \theta_{\text{c}}$, and a signature $\mathbf{m}$, CopyRNeRF~\cite{luo2023copyrnerf} builds the watermarked color representation relying on a number of MLPs to replace the original color in~\Eref{eq:nerffirst}. It first generates the color feature field and signature feature field as follows:
\begin{equation}
\mathbf{z}_{\text{c}}(\mathbf{x}, \mathbf{d})=g_{\text{c}}(\mathbf{c}(\mathbf{x}, \mathbf{d}),\gamma_{\mathbf{x}}(\mathbf{x}),\gamma_{\mathbf{d}}(\mathbf{d});\xi),
\label{eq:colorrepresentation}
\end{equation}
\begin{equation}
\mathbf{z}_{\text{m}}=g_{\text{m}}(\mathbf{m}; \phi),
\label{eq:messageembedding}
\end{equation}
where $\xi$ and $\phi$ are the parameters of the color feature encoder and signature feature encoder, respectively. Then, the watermarked color representation can be generated via a feature fusion module that integrates both the color feature field and the signature feature field as follows:
\begin{equation}
\mathbf{c}_{\text{m}}(\mathbf{x}, \mathbf{d})=g_{f}(\mathbf{z}_{\text{c}}(\mathbf{x}, \mathbf{d}),\mathbf{z}_{\text{m}}; \psi) + \mathbf{c}(\mathbf{x}, \mathbf{d}),
\label{eq:messagerepresentation}
\end{equation}
where $\psi$ is the parameter of the feature fusion module. In CopyRNeRF~\cite{luo2023copyrnerf}, the color feature encoder, signature feature encoder, and feature fusion module are all implemented using MLPs. The density in~\Eref{eq:sigma} is kept unchanged for better rendering quality. Finally, a CNN-based extractor is used to retrieve the predicted signature from the rendered image.

\section{Proposed Method}
\begin{figure*}
  \centering
  \includegraphics[width=\linewidth]{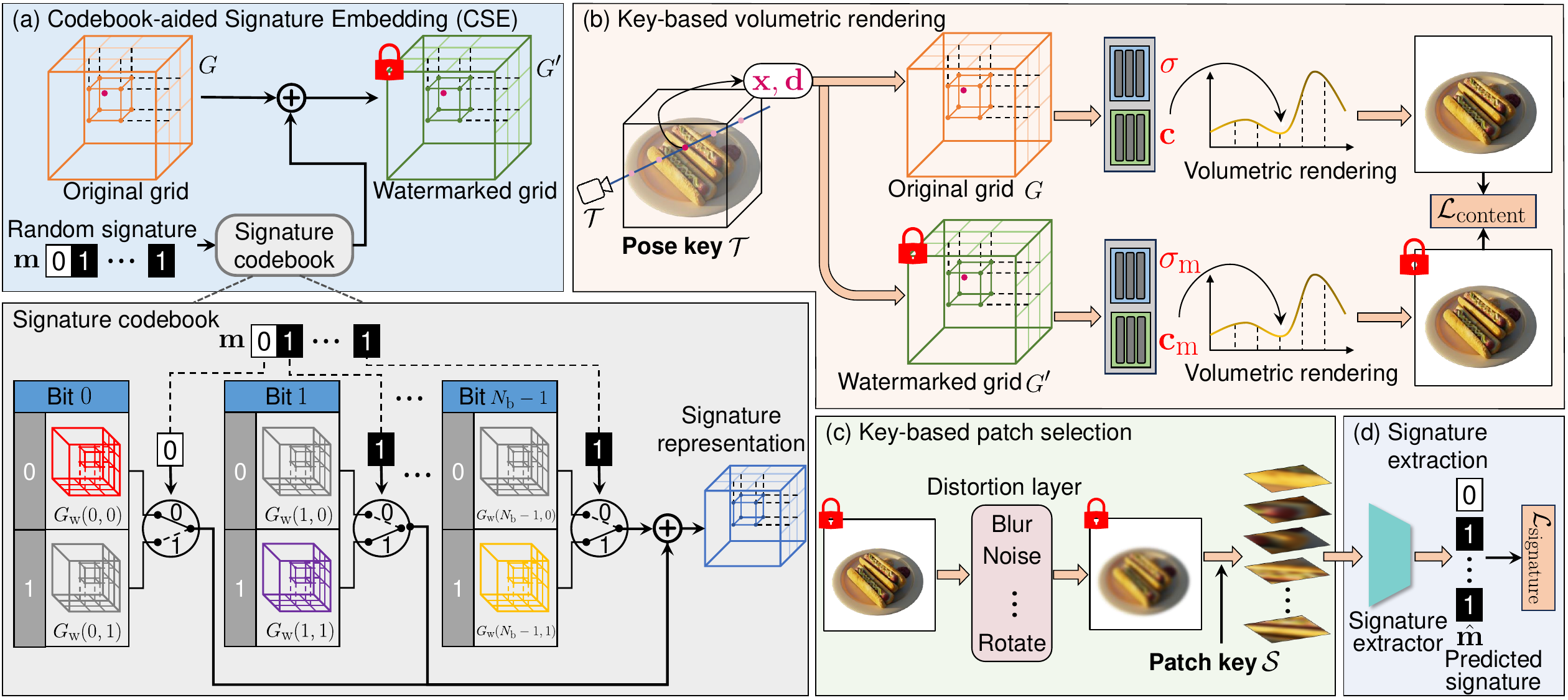}
  \caption{Illustration of our training pipeline.  (a) A signature representation $G_{\text{m}}$ is derived through a signature codebook according to the randomly selected signature $\mathbf{m}$. Subsequently, the watermarked grid $G^{\prime}$ is generated by directly adding the signature representation and original grid $G$ through a Codebook-aided Signature Embedding (CSE). (b) With a specific camera pose represented as pose key $\mathcal{T}$, we obtain an original image and a watermarked image by volumetric rendering from this pose, utilizing the original grid and the watermarked grid, respectively. A content loss $\mathcal{L}_{\text{content}}$ is computed by comparing the original image and watermarked image. (c) After a distortion layer, we use a patch key $\mathcal{S}$ to generate $N_{\text{b}}$ patches from the watermarked image. (d) We employ a signature extractor to extract one bit of signature from each patch. The signature loss $\mathcal{L}_{\text{signature}}$ is obtained by a cross-entropy error. The pose key $\mathcal{T}$ and the patch key $\mathcal{S}$ together form a complete key $\mathcal{K}=\{\mathcal{T}, \mathcal{S}\}$ for signature extraction.}

  \label{fig:framework}
\end{figure*}

As outlined in~\Fref{fig:framework}, we first construct a signature codebook to generate signature representation by representing each bit separately. Then, we incorporate the signature representation into the original NeRF by employing a straightforward addition operation through a Codebook-aided Signature Embedding (CSE), which results in a watermarked representation that retains the integrity of the original structure. During the verification, the signature can be extracted using a secret key, which indicates particular views and patches. Our method can achieve a high degree of imperceptibility and robustness at both the image level and model level. Additionally, the optimized signature codebook can take any binary signatures with length $N_{\text{b}}$ as the input to generate the corresponding signature representations, which allows the NeRF owner to flexibly embed any desired signatures in the NeRF.

\begin{figure}
  \centering
  \includegraphics[width=\linewidth]{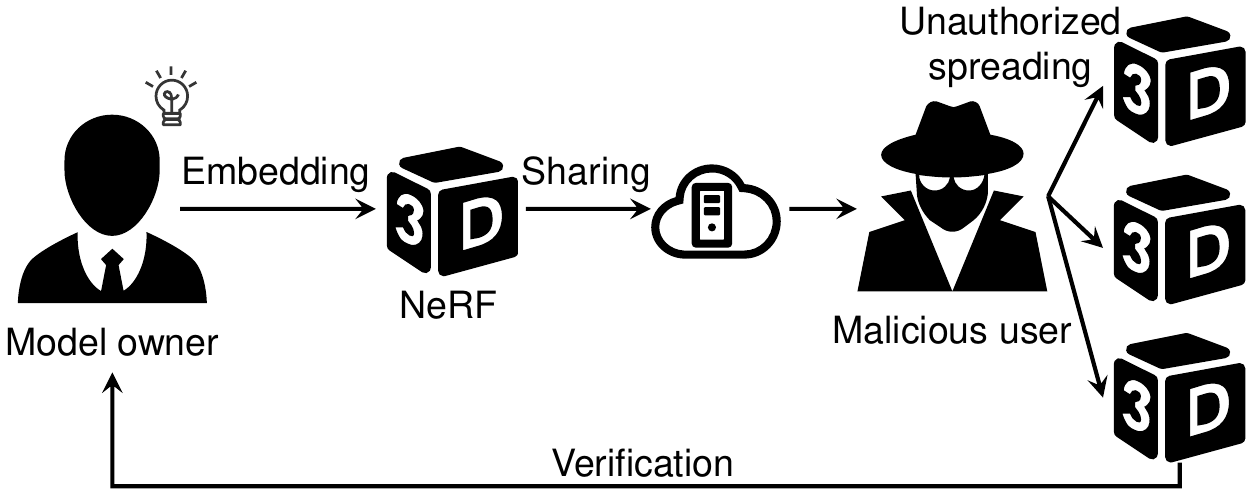}
  \caption{The threat model considered in our scenario. A NeRF owner generates the NeRF with signatures, and then the model is shared online. A malicious user obtains the shared model and spreads it without authorization. Finally, the NeRF owner can verify whether the model is generated by themselves.}  

  \label{fig:thread}
\end{figure}

\subsection{Threat model}
We first briefly describe the threat model considered in our work. NeRF watermarking aims to embed a specific signature into the NeRF without undermining the information within the NeRF, enabling the NeRF owner to identify if a given NeRF is embedded with this signature from the rendered images. Meanwhile, the malicious user may attempt to evade copyright verification by removing watermarks from the shared NeRF using image-level and model-level manipulations. As shown in~\Fref{fig:thread}, our threat model includes two agents, a NeRF owner and a malicious user, that act sequentially.
\begin{itemize}
    \item NeRF owner (embedding stage): The NeRF owner can generate protected NeRF through the watermarking algorithm. The watermarking algorithm should not degrade the quality of images rendered by NeRF and should not leave any visible marks.
    \item Malicious user: The malicious user obtains the shared NeRF with watermarks, then tries to evade the verification by applying manipulations on rendered images or NeRF parameter. Later, the malicious user uses the NeRF for a prohibited purpose and claims that the model is his/her intellectual property.
    \item NeRF owner (verification stage): The NeRF owner can extract signatures from their rendered images to verify whether the given NeRF is embedded with their signatures.
\end{itemize}

\subsection{Codebook-aided signature embedding}
\label{sec:mess}

We employ a signature codebook to generate a signature representation. The signature representation can be directly added to a portion of the NeRF parameters for signature embedding, which is called Codebook-aided Signature Embedding (CSE). This additive operation preserves the original NeRF structure while effectively embedding the watermark.

Specifically, we consider a given NeRF in a more general form:
\begin{equation}
[\sigma(\mathbf{x}), \mathbf{c}(\mathbf{x}, \mathbf{d})]=F\left(\mathbf{x}, \mathbf{d}; \Theta\right),
\label{eq:nerf}
\end{equation}
where $\sigma(\mathbf{x})$ and $\mathbf{c}(\mathbf{x}, \mathbf{d})$ are the density value and color value, respectively, and $\Theta$ is the whole parameters of the NeRF. We can split the parameters $\Theta$ into two portions. One portion is $\Theta_{\text{e}}$ for signature embedding, and the other portion is $\Theta_{\text{u}}$ kept unchanged. Therefore, the~\Eref{eq:nerf} can be rewritten as:
\begin{equation}
[\sigma(\mathbf{x}), \mathbf{c}(\mathbf{x}, \mathbf{d})]=F\left(\mathbf{x}, \mathbf{d}; \Theta_{\text{e}}, \Theta_{\text{u}}\right).
\label{eq:nerf_re}
\end{equation}

 Next, we use a signature codebook to generate the watermarking representation with the same structure as $\Theta_{\text{e}}$. Our basic idea is to represent each bit of the signature separately and then sum them up to obtain the final watermarked representation. 
 Therefore, this codebook can be designed with $2N_{\text{b}}$ representations to effectively encompass all $2^{N_{\text{b}}}$ potential signatures, where $N_{\text{b}}$ is the length of the binary signatures. 
This design of the learnable watermarking codebook ensures compact and efficient representation while maintaining flexibility for signature updates. The learnable structure is optimized during training to embed watermarks while maintaining rendering quality.

 In detail, we construct a learnable signature codebook $\mathcal{G}_{\text{w}}=\{G_{\text{w}}(n, k)\}_{n=0,k=0}^{N_{\text{b}}-1, 1}$. Each item in the $\mathcal{G}_{\text{w}}$ has the same structure as $\Theta_{\text{e}}$. For a specific binary signature $\mathbf{m}$, the $n$-th bit of $\mathbf{m}$ is denoted as $\mathbf{m}(n)$. The signature representation $G_{\text{m}}$ is constructed as:
\begin{equation}
G_{\text{m}}=\sum^{N_{\text{b}}-1}_{n=0}{G_{\text{w}}(n, \mathbf{m}(n))}.
\label{eq:w_grid}
\end{equation}
Then, the watermarked parameters can be obtained by adding the original parameters $\Theta_{\text{e}}$ to the signature representation $G_{\text{m}}$. We can use the watermarked parameters to generate the watermarked densities and colors as:
\begin{equation}
[\sigma_{\text{m}}(\mathbf{x}), \mathbf{c}_{\text{m}}(\mathbf{x}, \mathbf{d})]=F\left(\mathbf{x}, \mathbf{d}; \Theta_{\text{e}} + G_{\text{m}}, \Theta_{\text{u}}\right),
\label{eq:nerf_w}
\end{equation}
where $\sigma_{\text{m}}$ is the watermarked density and $\mathbf{c}_{\text{m}}$ is the watermarked color. This formulation preserves the structural integrity of the NeRF as the signature codebook matches the structure of $\Theta_{\text{e}}$, and the watermark is embedded through direct addition while keeping $\Theta_{\text{u}}$ unchanged.

The learnable signature codebook design offers several key advantages. First, it allows direct integration with NeRF parameters without requiring separate external modules, reducing potential vulnerabilities. Second, it achieves high efficiency in signature representation by accumulating individual bit representations. Third, NeRF owners can directly obtain new signature representations from the optimized codebook without additional finetuning.

The final 2D images are obtained through the standard volumetric rendering as follows:
\begin{equation}
\mathbf{C}_{\text{m}}(\mathbf{r})=\int_{t_n}^{t_f} T(t) \sigma_{\text{m}}(\mathbf{r}(t)) \mathbf{c}_{\text{m}}(\mathbf{r}(t), \mathbf{d}) d t,
\label{eq:water_rendering}
\end{equation}
where $T(t)$ is with the same definitions as their counterparts in~\Eref{eq:rendering2}.

As shown in~\Fref{fig:framework}, we illustrate an example of using Instant-NGP~\cite{muller2022instant} as the NeRF structure. Instant-NGP~\cite{muller2022instant} applies a grid-based data structure to speed up the training and rendering. We regard the grid parameters as $\Theta_{\text{e}}$ for signature embedding, and the other parameters are kept unchanged. More details can be found in~\Sref{sec:imp}.

Our method fully embeds the information into the NeRF parameters without using additional modules. Since the watermarking representation has the same structure as the original parameters $\Theta_{\text{e}}$, and they are integrated through addition, the structure of the NeRF remains unchanged. Therefore, malicious users cannot easily detect and remove the watermarks, which can guarantee higher model-level imperceptibility and robustness of our watermarking scheme.

\subsection{Joint pose-patch encryption watermarking}
Considering the extractor is unavoidable for signature extraction, we propose a joint pose-patch encryption watermarking strategy to apply it in a more encrypted manner for higher robustness. In our proposed approach, the signature is hidden in some patches from a particular perspective, with the camera pose and the patch locations regarded as the secret key. The NeRF owner should use this secret key to embed signatures in particular areas, and then use the same key to extract the signatures from these patches. Therefore, even when the extractor is accessed by malicious users, the signature should not be easily obtained by them because they are not aware of the actual key.

In our settings, one bit of signature can be extracted from each rendered patch. We express the camera pose as a camera-to-world transformation matrix $\mathcal{T}=[\mathbf{R}|\mathbf{t}] \in \mathrm{SE}(3)$, where $\mathbf{R} \in \mathrm{SO}(3)$ and $\mathbf{t} \in \mathbb{R}^3$ denote the camera rotation and translation, respectively. We refer to the pose key as the camera pose $\mathcal{T}$. We indicate the rendered patches from the camera pose as an ordered list $\mathcal{S}=\{(x_n, y_n)\}_{n=0}^{N_{\text{b}}-1}$, where $(x_n, y_n)$ is the coordinates of the center point for the $n$-th patch, and $N_{\text{b}}$ is length of the to-be-hidden binary signatures. We refer to the ordered list $\mathcal{S}$ as the patch key. Finally, we can represent the complete key by gathering the pose key and patch key as $\mathcal{K}=\{\mathcal{T}, \mathcal{S}\}$. With the known patch size of $h \times w$, $N_{\text{b}}$ patches can be uniquely determined through the key $\mathcal{K}$. 

With the secret key $\mathcal{K}$, the NeRF owner can use a regular CNN-based extractor to retrieve the signatures hidden in the patches. First, to render the $n$-th patch from the specific camera pose, we shoot $h\times w$ rays according to the pose key $\mathcal{T}$ and patch key $\mathcal{S}$. Then, we apply the volumetric rendering according to~\Eref{eq:water_rendering} to obtain each pixel color in the patch. Formally, we define a rendering operator $\mathcal{R}$ to obtain $N_{\text{b}}$ patches from the secret key as follows:
\begin{equation}
\mathcal{P} = \mathcal{R}(\Theta, \mathcal{K}),
\label{eq:render_op}
\end{equation}
where $\mathcal{P}=\{\mathbf{P}_n\}_{n=0}^{N_{\text{b}}-1}$ denotes the ordered list composed of $N_{\text{b}}$ rendered patches obtained from NeRF with parameter $\Theta$ using the secret key $\mathcal{K}$. Finally, the patches are fed into a CNN-based extractor to obtain the predicted signature as follows:
\begin{equation}
\hat{\mathbf{m}}(n) = f_{\text{m}}(\mathbf{P}_n; \theta_{\text{m}}),\quad n=0,1,\cdots, N_{\text{b}}-1,
\label{eq:extractor}
\end{equation}
where $\hat{\mathbf{m}}(n)$ indicates the $n$-th bit of the predicted signature $\hat{\mathbf{m}}$, and $\theta_{\text{m}}$ is the parameter of the CNN-based extractor. We can simplify the above equation as follows:
\begin{equation}
\hat{\mathbf{m}} = f_{\text{m}}(\mathcal{P}; \theta_{\text{m}}),
\label{eq:extractor_2}
\end{equation}
where $\mathcal{P}$ is the collection of all rendered patches.

\begin{figure}
  \centering
  \includegraphics[width=\linewidth]{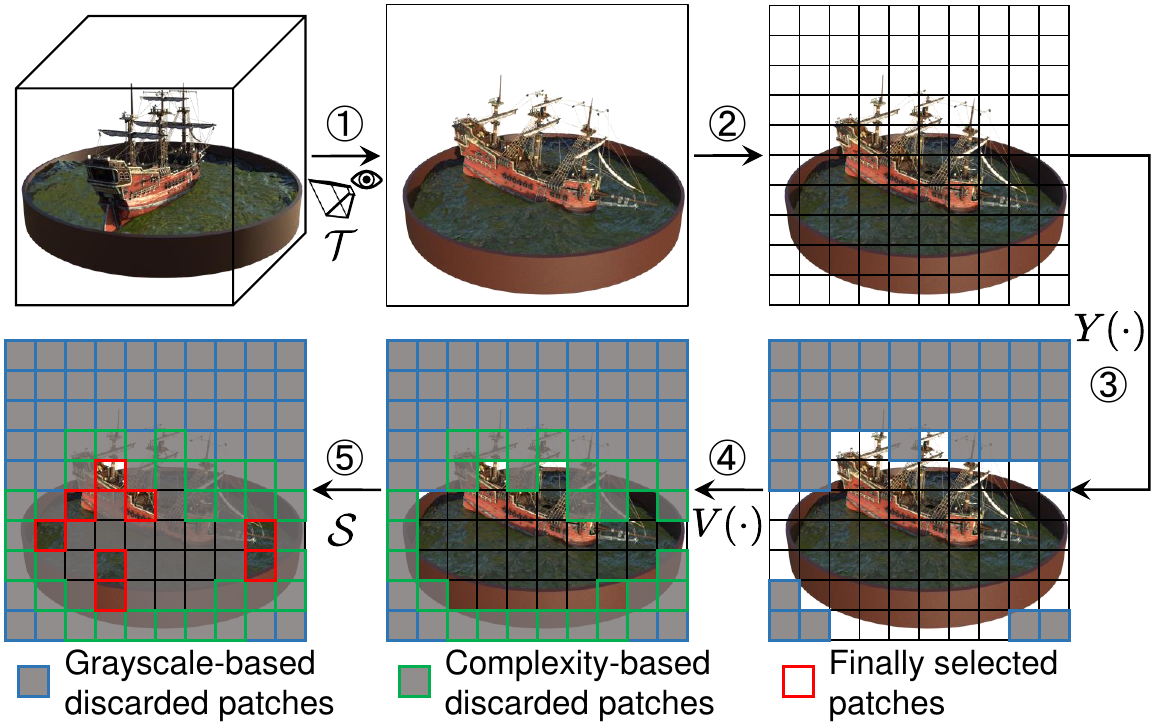}
  \caption{The CAKS scheme of our method. First, the NeRF owner chooses a camera pose as the pose key $\mathcal{T}$ and generates an image from this pose (\ding{192}). The rendered image is then uniformly divided into patches (\ding{193}). Subsequently, certain patches are discarded based on the grayscale values calculated by $Y(\cdot)$ (\ding{194}). Following that, patches with low complexity values are discarded using the complexity estimator $V(\cdot)$ (\ding{195}). Finally, the NeRF owner randomly selects $N_{\text{b}}$ patches as the final selection, and the positions of these patches form the patch key $\mathcal{S}$ (\ding{196}).}  

  \label{fig:patch}
\end{figure}
\subsection{Complexity-aware key selection scheme}
To further enhance the imperceptibility of watermarking, we propose a Complexity-Aware Key Selection (CAKS) scheme, as shown in~\Fref{fig:patch}. The embedded regions have been proven to play an important role in information hiding~\cite{pan2021seek}, with optimal patches offering superior imperceptibility for watermarking~\cite{pan2021seek, filler2010gibbs}. In general, areas with more complex textures are more optimal for information hiding~\cite{meng2018fusion, huang2019image}. Therefore, our CAKS scheme can select patches with high visual complexity values for embedding watermarks.

In detail, we first randomly select a camera pose $\mathcal{T}$ as a pose key. From the camera poses $\mathcal{T}$, we can obtain the rendered images $\mathbf{I}$, which has a full size of $H \times W$. Then, we evenly partition each rendered image $\mathbf{I}$ into patches, each with the size of $h \times w$. These patches can be indicated by the coordinates of center points as a list $\mathcal{S}_0 = \{(x_i, y_j)\}_{i=0,j=0}^{N_{\text{h}}-1,N_{\text{w}}-1}$: 
\begin{equation}
\begin{cases}
x_i = i\times h + \frac{h}{2},& i=0,1,\cdots, N_{\text{h}}-1,\\
y_j = j\times w + \frac{w}{2},& j=0,1,\cdots, N_{\text{w}}-1,
\label{eq:patch}
\end{cases}
\end{equation}
where $N_{\text{h}} = \lfloor{\frac{H}{h}}\rfloor$ and $N_{\text{w}} = \lfloor{\frac{W}{w}}\rfloor$ are the number of divisions along the height and width, respectively. According to the coordinates of center points and patch size, the coordinates of all $h\times w$ points in each patch can be easily obtained. Thus, all patches can be rendered and assembled into $\mathcal{P}_0=\{\mathbf{P}_{i,j}\}_{i=0,j=0}^{N_{\text{h}}-1,N_{\text{w}}-1}$, where $\mathbf{P}_{i,j}$ is the rendered RGB patch centered at $(x_i, y_j)$ from the camera pose $\mathcal{T}$. For convenience, the patch list can be rewritten as $\mathcal{P}_0 = \{\mathbf{P}_n\}_{n=0}^{N_{\text{h}} \times N_{\text{w}} - 1}$ with a single subscript.

However, not all the patches are suitable for watermarking. It has been proven that hiding information in areas with low color variations can easily leave detectable traces~\cite{luo2023copyrnerf}, compromising the imperceptibility of the watermarking on rendered images. One scenario where these areas appear is in the rendering backgrounds of some object-only 3D models, with many 3D assets falling into this important category~\cite{jain2022zero}. Therefore, we use a simple method to filter out these patches. Specifically, we calculate the color variation within patch $\mathbf{P}$ as follows:
\begin{equation}
Y(\mathbf{P}) = \frac{1}{3}(\texttt{var}(\mathbf{P}^{\text{R}})+\texttt{var}(\mathbf{P}^{\text{G}})+\texttt{var}(\mathbf{P}^{\text{B}})),
\label{eq:var_def}
\end{equation}
where \texttt{var} represents the variance of the rendered patch in a specific channel, and $\mathbf{P}^{\text{R}}$, $\mathbf{P}^{\text{G}}$, $\mathbf{P}^{\text{B}}$ represent the three channels of the patch, respectively. By setting a variance threshold $\delta_{\text{var}}$, the candidate patches $\mathcal{P}_1$ can be obtained as follows:
\begin{equation}
\mathcal{P}_1 = \{\mathbf{P}| \mathbf{P} \in \mathcal{P}_0, Y(\mathbf{P}) < \delta_{\text{var}}\},
\label{eq:gray_th}
\end{equation}
where $Y(\cdot)$ is defined in~\Eref{eq:var_def}, and $\delta_{\text{var}}$ is the threshold to remove patches with lower variance values.

Next, we select patches with high visual complexity from the candidate patches $\mathcal{P}_1$ to form the final list of candidate patches $\mathcal{P}_2$. We use a well-established visual complexity estimator $V(\cdot)$ to predict the complexity value of each patch. Specifically, the complexity estimator calculates the visual complexity as the ratio between the compressed and uncompressed image storage size as follows~\cite{donderi2006visual, forsythe2011predicting}:
\begin{equation}
V(\mathbf{P}) = \frac{\texttt{Size}(\texttt{Compress}(\mathbf{P}))}{\texttt{Size}(\mathbf{P})},
\label{eq:complexity}
\end{equation}
where $\texttt{Size}(\mathbf{P})$ is the storage size of the uncompressed patch $\mathbf{P}$, and $\texttt{Size}(\texttt{Compress}(\mathbf{P}))$ is storage size of the output of compressor $\texttt{Compress}(\cdot)$.

Patches with low complexity values are discarded, and the NeRF owner can randomly select $N_{\text{m}}$ patches. The patch key $\mathcal{S}$ is then generated according to the locations of these patches. To this end, we propose to use an approach for key selection. After calculating the complexity of each patch in $\mathcal{P}_1$, we can obtain the candidate patches $\mathcal{P}_2$ as follows:
\begin{equation}
\mathcal{P}_2 = \{\mathbf{P}| \mathbf{P} \in \mathcal{P}_1, V(\mathbf{P}) > \delta_{\text{complexity}}\},
\label{eq:comp_th}
\end{equation}
where $\delta_{\text{complexity}}$ is a threshold for complexity values.

The last step for the NeRF owner is to randomly select $N_{\text{b}}$ patches out of $\mathcal{P}_2$ for signature embedding and extraction. The center point coordinates of these $N_{\text{b}}$ patches form the patch key $\mathcal{S}$. 

\begin{figure*}[htbp]
  \centering
  \includegraphics[width=\linewidth]{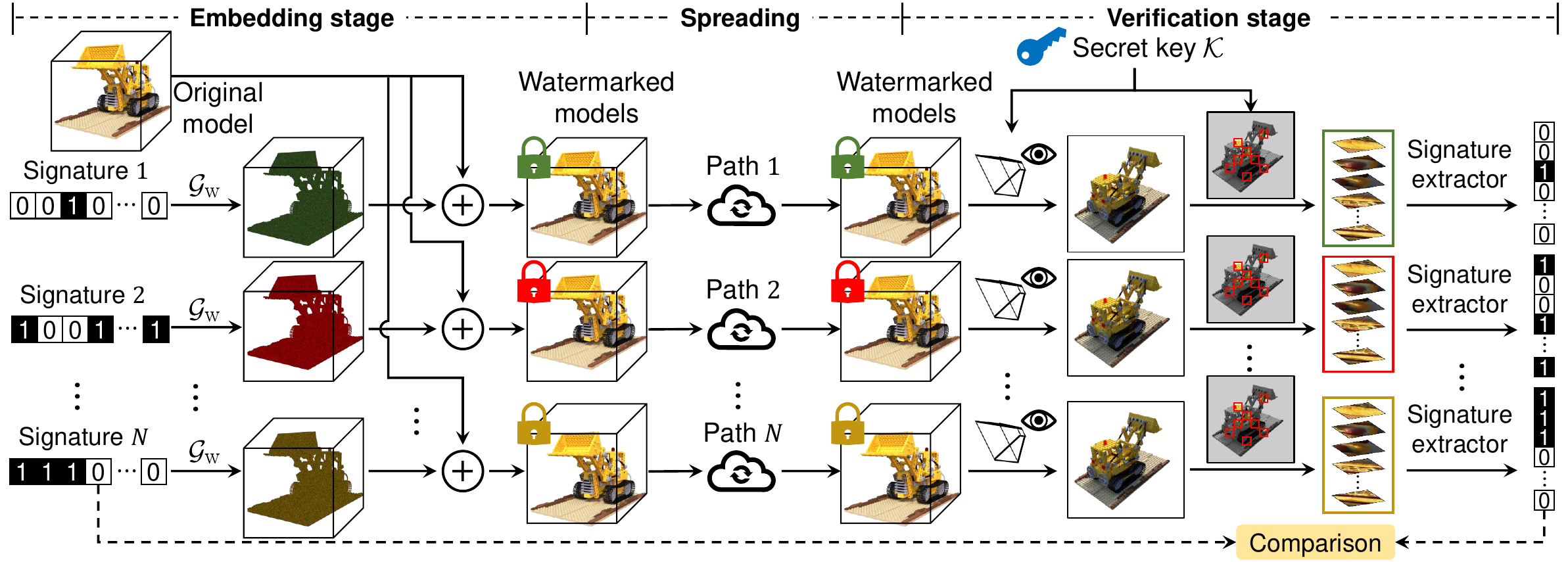}
  \caption{Illustration of our workflow when the signature codebook $\mathcal{G}_{\text{w}}$ has been optimized. In the embedding stage, the NeRF owner can select $N$ signatures to embed into the to-be-protected model through the CSE, resulting in $N$ watermarked models with different embedded signatures. These models are shared online through different paths. In the verification stage, the NeRF owner first obtains specific patches by rendering the model through a key $\mathcal{K}$ for perspective and patch selection for the to-be-verified model. Then, an extractor extracts the signatures embedded in different to-be-verified models. By comparing these signatures with the originally embedded signatures, the NeRF owner can determine model ownership and trace the path through which the model is abused.}
 
  \label{fig:inference}
\end{figure*}

\subsection{Distortion layer}
The robustness is a unique part that makes digital watermarking different from other information-hiding tasks~\cite{zhu2018hidden, wang2023data}. We follow our settings in CopyRNeRF to use a Distortion Layer (DiL) module to enhance the robustness of watermarks against image transformations. The DiL module is positioned after volumetric rendering but before patch selection, as shown in~\Fref{fig:framework}. During optimization, the DiL module can simulate degradation rendered images might encounter, such as blur noise. The DiL module is discarded after optimization. Our watermarking method can ensure that the embedded signatures within the NeRF can be precisely extracted by the extractor, even when the images rendered from the NeRF are subjected to a range of distortions. In our experiments, we demonstrate the DiL module's effectiveness in enhancing the system's robustness.

\subsection{Optimization}
In our setting, we aim to ensure both the imperceptibility of the embedding process and the successful extraction of signatures. We jointly train the signature codebook $\mathcal{G}_{\text{w}}$ and extractor $\theta_{\text{m}}$ end-to-end. 

We construct a content loss $\mathcal{L}_{\text{content}}$ to ensure the imperceptibility. In detail, the content loss $\mathcal{L}_{\text{content}}$ is obtained by computing the MSE as follows:
\begin{equation}
  \mathcal{L}_{\text{content}} = \mathbb{E}_{\mathbf{r} \in \mathcal{B}}\|\mathbf{C}_{\text{m}}(\mathbf{r})-\mathbf{C}(\mathbf{r})\|_2^2,
  \label{eq:content}
\end{equation}
where $\mathcal{B}$ is the set of rays in a batch, $\mathbf{C}_{\text{m}}(\mathbf{r})$ and $\mathbf{C}(\mathbf{r})$ are rendered from watermarked grid as~\Eref{eq:water_rendering} and original grid as~\Eref{eq:rendering}, respectively. 

We randomly choose a binary signature within one optimization loop to enable the NeRF owner to conveniently select any secret signatures for embedding after optimization. In detail, we randomly select a binary signature $\mathbf{m}$ of length $N_{\text{b}}$. Then, $N_{\text{b}}$ bits can be obtained through~\Eref{eq:extractor} as predicted signature $\hat{\mathbf{m}}$. The signature loss $\mathcal{L}_{\text{signature}}$ is finally obtained by calculating the masked binary cross-entropy error between predicted signature $\hat{\mathbf{m}}$ and the ground truth signature $\mathbf{m}$ as follows:
\begin{equation}
\begin{split}
      \mathcal{L}_{\text{signature}} = \frac{1}{N_{\text{b}}} \sum_{n=0}^{N_{\text{b}}-1} &-[\mathbf{m}(n) \log \hat{\mathbf{m}}(n)+\\ &(1-\mathbf{m}(n)) \log (1-\hat{\mathbf{m}}(n))],
\end{split}
\label{eq:loss_m}
\end{equation}
where $\mathbf{m}(n)$ and $\hat{\mathbf{m}}(n)$ indicate the $n$-th bit of the ground truth signature and predicted signature, respectively.

Therefore, the overall loss $\mathcal{L}_{\text{overall}}$ can be obtained as:
\begin{equation}
\mathcal{L}_{\text{overall}} = \mathcal{L}_{\text{content}} + \gamma_{\text{signature}}\mathcal{L}_{\text{signature}},
\label{eq:loss}
\end{equation}
where $\gamma_{\text{signature}}$ is the hyperparameter to balance the loss functions.

\subsection{Workflow after optimization}
\label{sec:inf}

After optimization of the signature codebook, the NeRF owner can directly watermark any NeRF with arbitrary signatures of length $N_{\text{b}}$ through a simple parameter addition, requiring no model retraining or fine-tuning. As shown in~\Fref{fig:inference}, during the embedding stage, the NeRF owner first randomly chooses $N$ secret binary signatures with length $N_{\text{b}}$. These signatures are then efficiently embedded into the NeRF through direct parameter addition to obtain $N$ watermarked NeRFs. This process is computationally efficient as it requires only straightforward addition operations ($\Theta_{\text{e}} + G_{\text{m}}$) without any training overhead. Each watermarked model can be shared through a different path (\ie, different distribution channel such as website, platform, or user group), enabling the tracking of potential model leaks or misuse. Specifically, the watermarked models are obtained through~\Eref{eq:w_grid} and~\Eref{eq:nerf_w} based on the optimized signature codebook $\mathcal{G}_{\text{w}}$. Subsequently, the watermarked models can be shared with the public. The NeRF owner should keep the secret key $\mathcal{K}$ determined during optimization.

During the verification, the NeRF owner has several to-be-verified models. The NeRF owner can render specific patches according to the secret key $\mathcal{K}$ by querying these models, and an extractor is applied to obtain the predicted signatures $\hat{\mathbf{m}}$ through~\Eref{eq:extractor}. By comparing the predicted signatures to the originally embedded signatures, the owner can determine which models belong to them. As models embedded with different signatures spread through their respective paths, the NeRF owner can trace which paths the models are abused through. To evaluate the bit accuracy during the verification stage, the binary predicted signature $\hat{\mathbf{m}}_{\text{b}}$ can be obtained by rounding:
\begin{equation}
\hat{\mathbf{m}}_{\text{b}} =  {\texttt{clamp}}({\texttt{sign}}(\hat{\mathbf{m}}), 0, 1),
\label{eq:bit_m}
\end{equation}
where \texttt{clamp} and \texttt{sign} are of the same definitions in~\cite{Yoo_2022_CVPR}. It should be noted that we use the continuous result $\hat{\mathbf{m}}$ to compute loss in the training process, while the binary one $\hat{\mathbf{m}}_{\text{b}}$ is only adopted after optimization to compute bit accuracy.

\subsection{Implementation details}
\label{sec:imp}
We implement our method using PyTorch. The to-be-protected models are obtained using Instant-NGP~\cite{muller2022instant}, a popular NeRF model, with the suggested settings in the original paper~\cite{muller2022instant}. Due to the human visual system's reduced sensitivity to high-frequency details, the signature codebook $\mathcal{G}_{\text{w}}$ is constructed only to modify the grid at the finest resolution. In such a setting, $\Theta_{\text{e}}$ in~\Eref{eq:nerf_re} refers to the parameters of the grid at the finest resolution, and $\Theta_{\text{e}}$ refers to the other parameters of NeRF. Embedding watermarks in high frequencies allows for more covert hiding of watermark information, minimizing its perceptual impact on the original media. The extractor consists of $7$ blocks, where each block is composed of 2D convolutional layers with batch normalization and ReLU activation functions. This is followed by a block with the desired output dimension, the signature length hidden in each patch, an average pooling layer, and a final linear layer. The number of divisions along the height $N_{\text{h}}$ and width $N_{\text{w}}$ of one rendered image in~\Eref{eq:patch} is set as $32$. The $\delta_{\text{gray}}$ in~\Eref{eq:gray_th} is set as $0.9$. The $\delta_{\text{complexity}}$ in~\Eref{eq:comp_th} is set to control the size of candidate patches $\mathcal{P}_2$ as $1.5N_{\text{b}}$. The views are randomly selected to embed and extract the watermarking signature. The hyperparameter in~\Eref{eq:loss} is set as $\gamma_{\text{signature}}=10.0$. We use the Adam optimizer with default values $\beta_1 = 0.9$, $\beta_2 = 0.999$, $\epsilon = 10^{-8}$, and an initial learning rate $1\times10^{-3}$ that decays following the exponential scheduler with weight decay $5\times10^{-4}$ during optimization. We jointly train the signature codebook $\mathcal{G}_{\text{w}}$, and extractor $\theta_{\text{m}}$ for $3K$ iterations on a single NVIDIA Tesla V100 GPU.

\begin{figure*}
  \centering
  \includegraphics[width=\linewidth]{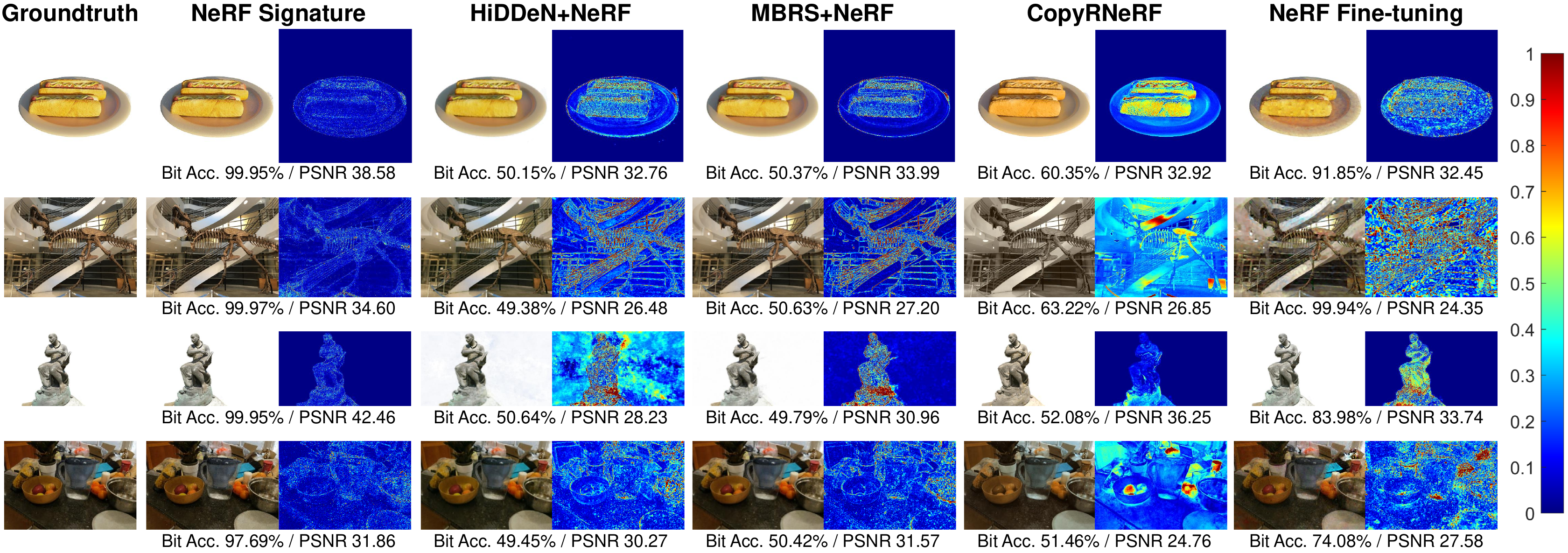}
  \caption{
      Visual quality compared with the baselines. Results are presented for $48$ bits. For each method, we show the original rendering (left) and the corresponding difference map (right). The difference maps visualize the pixel-wise residuals between watermarked and original renderings with 10$\times$ amplification for better visibility. The averaged Bit Acc. / PSNR is shown below each example. From top to bottom: ``hotdog'' from Blender, ``trex'' from LLFF, ``Ignatius'' from Tanks \& Temples, and ``counter'' from Mip-NeRF360.
  }
  \label{fig:result}
\end{figure*} 

\section{Experiments}

\subsection{Experimental setting}
\noindent{\textbf{Datasets.}} We evaluate our methods on $4$ established datasets, including Blender dataset~\cite{nerf2020}, LLFF dataset~\cite{mildenhall2019local}, Tanks \& Temples dataset~\cite{knapitsch2017tanks}, and Mip-NeRF360 dataset~\cite{barron2022mip}, which are commonly used datasets for novel view synthesis. The Blender dataset contains $8$ detailed synthetic objects with $100$ images from virtual cameras arranged on a hemisphere pointed inward. LLFF dataset consists of $8$ real-world scenes that contain mainly forward-facing images. Each scene contains images of $20$ to $62$. Tanks \& Temples dataset~\cite{knapitsch2017tanks} contains realistic scenes with large central objects, including Caterpillar, Family, Truck, \etc~Each scene within this dataset consists of $263$ to $1107$ images captured using a monocular RGB camera. Mip-NeRF360 dataset consists of $9$ indoor and outdoor scenes with complex central objects. We follow the settings in the Instant-NGP~\cite{muller2022instant} to obtain the reconstruction result to be protected. Based on the obtained model, our watermarking method is implemented. 
Due to the need to embed information from random perspectives, we directly use the rendered images of the to-be-protected model from specific perspectives as references instead of using original images to train the to-be-protected model.

\noindent{\textbf{Baselines.}}
We compare our method with four deep watermarking models to guarantee a fair comparison: 1) \textbf{HiDDeN~\cite{zhu2018hidden}+NeRF~\cite{muller2022instant}}: processing images from different viewing directions with a classical 2D watermarking method HiDDeN~\cite{zhu2018hidden} before training the NeRF; 2) \textbf{MBRS~\cite{jia2021mbrs}+NeRF~\cite{muller2022instant}}: processing images with state-of-the-art 2D watermarking method MBRS~\cite{jia2021mbrs} before training the NeRF representation; 3) \textbf{CopyRNeRF~\cite{luo2023copyrnerf}}: a method to embed the secret signature into the NeRF with model structure changed; 4) \textbf{NeRF Fine-tuning}: a method to watermark the NeRF by fine-tuning the model every time the NeRF owner wants to embed a different signature. All the above baselines use the same NeRF structure as our method, Instant-NGP~\cite{muller2022instant}, to ensure fairness in comparison.

\noindent{\textbf{Evaluation methodology.}}  
We evaluate our NeRF Signature across three key aspects: accuracy, imperceptibility, and robustness. For \textit{accuracy}, we measure the bit accuracy, defined as the percentage of correctly extracted bits, by comparing the binary predicted signature obtained from~\Eref{eq:bit_m} with the ground truth binary signature. The bit accuracy is computed as the average over $200$ randomly selected binary signatures. For \textit{imperceptibility}, we assess the impact of digital watermarking on the original to-be-protected model. To mitigate the influence of the original NeRF performance, we compare the watermarked model with its corresponding original model, as this relative comparison isolates the impact of the watermarking process and provides a precise measure of imperceptibility without being influenced by the absolute quality of the original NeRF. Image-level imperceptibility is evaluated using Peak Signal-to-Noise Ratio (PSNR), Structural Similarity Index (SSIM)~\cite{wang2004image}, and Learned Perceptual Image Patch Similarity (LPIPS)~\cite{zhang2018unreasonable}, which measure the visual quality of rendered images after signature embedding. For each scene, $360$ rendered images from different viewpoints are used for evaluation. Additionally, \textit{model-level imperceptibility} is ensured by preserving the structure of the watermarked model identical to the original model, as discussed in~\Sref{sec:mess}. For \textit{robustness}, we test the ability to extract embedded signatures under image-level transformations and model-level modifications. We consider $7$ types of image-level transformations, including rotation, cropping, scaling, JPEG compression, blurring, noise, and brightness adjustments. For model-level modifications, we focus on fine-tuning and adversarial attacks. Adversarial attacks are implemented using the Projected Gradient Descent (PGD) method~\cite{madry2018towards} to generate adversarial examples of NeRF that aim to fool the signature extractor. Furthermore, we evaluate robustness against private key attacks by testing whether malicious users can uncover the signature using guessed keys after obtaining the extractor.

\begin{table*}[htbp]
  \centering
  \caption{Bit accuracy and imperceptibility compared with the baselines. $\uparrow$ ($\downarrow$) indicates higher (lower) metric value is better. Results are presented for $16$, $32$, and $48$ bits and are averaged across all instances within each dataset. The best-performing methods are highlighted in \textbf{bold}.}
    \resizebox{\textwidth}{!}{
    \begin{tabular}{c|c|cccc|cccc|cccc}
    \hline
    \multirow{2}[3]{*}{Dataset} & \multirow{2}[3]{*}{Method} & \multicolumn{4}{c|}{16 bit}   & \multicolumn{4}{c|}{32 bit}   & \multicolumn{4}{c}{48 bit} \bigstrut[t]\\
\cline{3-14}          &       & Bit Acc.$\uparrow$ & PSNR$\uparrow$ & SSIM$\uparrow$ & LPIPS$\downarrow$ & Bit Acc.$\uparrow$ & PSNR$\uparrow$ & SSIM$\uparrow$ & LPIPS$\downarrow$ & Bit Acc.$\uparrow$ & PSNR$\uparrow$ & SSIM$\uparrow$ & LPIPS$\downarrow$ \bigstrut[t]\\
    \hline
    \multirow{5}[0]{*}{\rotatebox{90}{Blender}} & NeRF Signature & \textbf{99.98\%} & \textbf{43.31} & \textbf{0.9949} & \textbf{0.0028} & \textbf{99.98\%} & \textbf{36.18} & \textbf{0.9796} & \textbf{0.0149} & \textbf{99.94\%} & \textbf{31.80} & \textbf{0.9628} & \textbf{0.0323} \bigstrut[t]\\
          & HiDDeN~\cite{zhu2018hidden}+NeRF & 50.38\% & 31.11 & 0.9451 & 0.0482 & 49.59\% & 27.31 & 0.9319 & 0.0695 & 50.29\% & 25.89 & 0.9302 & 0.0780 \\
          & MBRS~\cite{jia2021mbrs}+NeRF & 51.78\% & 32.95 & 0.9791 & 0.0292 & 50.98\% & 28.25 & 0.9513 & 0.0577 & 50.53\% & 26.78 & 0.9402 & 0.0697 \\
          & CopyRNeRF~\cite{luo2023copyrnerf} & 90.32\% & 27.31 & 0.9373 & 0.0562 & 79.22\% & 24.50 & 0.9268 & 0.0720 & 62.15\% & 23.76 & 0.9135 & 0.0799 \\
          & NeRF Fine-tuning & 94.09\% & 33.69 & 0.9828 & 0.0215 & 87.19\% & 29.12 & 0.9402 & 0.0467 & 85.26\% & 27.06 & 0.9310 & 0.0513 \\
    \hline
    \multirow{5}[0]{*}{\rotatebox{90}{LLFF}} & NeRF Signature & \textbf{99.99\%} & \textbf{34.09} & \textbf{0.9555} & \textbf{0.0365} & \textbf{99.94\%} & \textbf{31.27} & \textbf{0.9037} & \textbf{0.1072} & \textbf{99.48\%} & \textbf{29.88} & \textbf{0.8943} & \textbf{0.1188} \bigstrut[t]\\
          & HiDDeN~\cite{zhu2018hidden}+NeRF& 52.71\% & 31.39 & 0.8843 & 0.1196 & 48.13\% & 27.21 & 0.8423 & 0.1334 & 51.04\% & 26.25 & 0.8373 & 0.1436 \\
          & MBRS~\cite{jia2021mbrs}+NeRF & 51.25\% & 31.78 & 0.9155 & 0.1029 & 50.42\% & 28.14 & 0.8654 & 0.1192 & 48.33\% & 27.66 & 0.8523 & 0.1223 \\
          & CopyRNeRF~\cite{luo2023copyrnerf} & 85.45\% & 26.53 & 0.9023 & 0.1112 & 74.36\% & 24.96 & 0.8914 & 0.1238 & 56.35\% & 23.51 & 0.8894 & 0.1343 \\
          & NeRF Fine-tuning & 97.29\% & 27.82 & 0.7958 & 0.2059 & 91.24\% & 25.70 & 0.7447 & 0.2389 & 86.31\% & 23.68 & 0.7345 & 0.2546 \\
    \hline
    \multirow{5}[1]{*}{\rotatebox{90}{\begin{tabular}[c]{@{}c@{}}Tanks \&\\ Temples\end{tabular}}} & NeRF Signature & \textbf{100.00\%} & \textbf{44.45} & \textbf{0.9884} & \textbf{0.0080} & \textbf{99.98\%} & \textbf{38.10} & \textbf{0.9564} & \textbf{0.0183} & \textbf{99.92\%} & \textbf{34.06} & \textbf{0.9231} & \textbf{0.0457} \bigstrut[t]\\
          & HiDDeN~\cite{zhu2018hidden}+NeRF & 52.92\% & 29.54 & 0.9186 & 0.0351 & 50.83\% & 26.14 & 0.8932 & 0.0596 & 49.38\% & 25.07 & 0.8806 & 0.0619 \\
          & MBRS~\cite{jia2021mbrs}+NeRF & 52.08\% & 31.18 & 0.9347 & 0.0257 & 50.21\% & 27.36 & 0.9096 & 0.0436 & 50.42\% & 26.70 & 0.8962 & 0.0545 \\
          & CopyRNeRF~\cite{luo2023copyrnerf} & 86.24\% & 35.47 & 0.9678 & 0.0095 & 71.58\% & 31.36 & 0.9353 & 0.0254 & 55.96\% & 29.56 & 0.9173 & 0.0477 \\
          & NeRF Fine-tuning & 92.50\% & 33.57 & 0.9496 & 0.0124 & 87.80\% & 30.32 & 0.9188 & 0.0358 & 83.10\% & 28.98 & 0.9001 & 0.0503 \\
    \hline
    \multirow{5}[2]{*}{\rotatebox{90}{\begin{tabular}[c]{@{}c@{}}Mip-\\ NeRF360\end{tabular}}} & NeRF Signature & \textbf{99.97\%} & \textbf{33.07} & \textbf{0.9223} & 0.0583 & \textbf{97.39\%} & \textbf{31.12} & \textbf{0.9112} & 0.1013 & \textbf{96.47\%} & \textbf{30.68} & \textbf{0.8948} & 0.1113 \bigstrut[t]\\
          & HiDDeN~\cite{zhu2018hidden}+NeRF & 52.71\% & 32.76 & 0.8876 & 0.0629 & 51.67\% & 30.92 & 0.8532 & 0.0980 & 49.79\% & 29.54 & 0.8482 & 0.1018 \\
          & MBRS~\cite{jia2021mbrs}+NeRF & 52.50\% & 32.82 & 0.9095 & \textbf{0.0526} & 52.08\% & 31.09 & 0.8753 & \textbf{0.0836} & 50.63\% & 30.57 & 0.8634 & \textbf{0.0942} \\
          & CopyRNeRF~\cite{luo2023copyrnerf} & 79.42\% & 29.25 & 0.8598 & 0.1071 & 68.81\% & 27.98 & 0.8233 & 0.1224 & 53.16\% & 26.45 & 0.8108 & 0.1373 \\
          & NeRF Fine-tuning & 91.25\% & 30.36 & 0.8742 & 0.0976 & 84.88\% & 28.84 & 0.8414 & 0.1129 & 80.50\% & 27.59 & 0.8366 & 0.1214 \\
    \hline
    \end{tabular}%
    }
  \label{tab:main_results}%
\end{table*}%

\subsection{Accuracy and imperceptibility}
Accuracy and imperceptibility are two important indicators of digital watermarking. We calculate the bit accuracy by comparing the extracted binary signature with the embedded one. We randomly select multiple binary signatures and obtain the final bit accuracy by averaging. We compare the rendered images from watermarked NeRF and the original NeRF to evaluate imperceptibility. We report the final imperceptibility using the average metrics from multiple viewpoints.
\subsubsection{Qualitative results}

We first evaluate the imperceptibility visually compared to all baselines, with the results presented in~\Fref{fig:result}. To quantitatively visualize the watermarking artifacts, we compute pixel-wise residual maps between the watermarked and original renderings, with an amplification factor of $10$. The bit accuracy of each method is also shown in the results.
 Due to the effectiveness of our NeRF Signature, our method can achieve a high level of imperceptibility by keeping the rendered images changed little from the original content. Besides, our method can also accurately extract the embedded signature with bit accuracy $100\%$. Although HiDDeN~\cite{zhu2018hidden}+NeRF~\cite{muller2022instant} and MBRS~\cite{jia2021mbrs}+NeRF~\cite{muller2022instant} both yield high-quality reconstructions, their bit accuracy values for the rendered images are low, near a random guess probability of $50\%$. This indicates that the signatures cannot be efficiently embedded after the training of the NeRF~\cite{luo2023copyrnerf}. CopyRNeRF~\cite{luo2023copyrnerf} reaches a balance between accuracy and imperceptibility, but some degradation is visually perceptible in the rendered images. Moreover, the bit accuracy is low when the bit length is long. NeRF Fine-tuning can achieve good visual quality, but each fine-tuning session can only embed a fixed signature into the model. Whenever the NeRF owners need to change the embedded signature, they must repeatedly fine-tune the NeRF, which undermines flexibility for users.

\subsubsection{Quantitative results}
We present the quantitative bit accuracy and imperceptibility results across various bit length settings in~\Tref{tab:main_results}. NeRF Signature achieves the highest bit accuracy across all experimental bit lengths. Moreover, high PSNR and SSIM values, along with low LPIPS values, indicate that the images rendered from the watermarked model by our method can preserve the visual fidelity and structural integrity of the original content, ensuring minimal perceptual distortion. The results from HiDDeN~\cite{zhu2018hidden}+NeRF~\cite{muller2022instant} and MBRS~\cite{jia2021mbrs}+NeRF~\cite{muller2022instant} demonstrate that they can achieve a high level of imperceptibility, but they are unable to maintain a high bit extraction accuracy, even when the bit length is short. Although CopyRNeRF~\cite{luo2023copyrnerf} can accurately extract signatures for bit length $8$, we can see that the bit accuracy drops when the number of bits increases. NeRF Fine-tuning can reach the second-best results among all settings, which shows that such an approach can also effectively embed signatures into NeRF. However, the NeRF Fine-tuning method still encounters the issue of repeated fine-tuning when the hidden signatures need to be changed.

To provide a comprehensive evaluation of absolute rendering quality, we conduct an experiment by comparing all rendered images against the ground truth images captured from the real world. The results are shown in~\Tref{tab:rebuttal}. The first row is the results for non-watermark NeRF, which represents the original model quality. The results show that our NeRF Signature maintains rendering quality closest to the original model while achieving significantly higher bit accuracy than other watermarking methods.

\begin{table}[htbp]
  \centering
  
  \caption{Comparison of different methods on quality metrics and bit accuracy for embedding 48-bit watermark on ``Ignatius" scene from Tanks \& Temples dataset~\cite{knapitsch2017tanks}. $\uparrow$ ($\downarrow$) indicates higher (lower) metric value is better.}
    % \resizebox{\linewidth}{!}
    {
    \begin{tabular}{c|cccc}
    \hline
    Method & PSNR$\uparrow$ & SSIM$\uparrow$ & LPIPS$\downarrow$ & Bit Acc.$\uparrow$ \bigstrut[t]\\
    \hline
    non-watermark & {28.12} & {0.953} & {0.048} & - \bigstrut[t]\\
    NeRF Signature & 27.95 & 0.948 & 0.051 & {99.95\%} \bigstrut[t]\\
    HiDDeN~\cite{zhu2018hidden}+NeRF & 26.68 & 0.924 & 0.178 & 50.64\% \bigstrut[t]\\
    MBRS~\cite{jia2021mbrs}+NeRF & 27.67 & 0.945 & 0.075 & 49.79\% \bigstrut[t]\\
    CopyRNeRF~\cite{luo2023copyrnerf} & 27.47 & 0.944 & 0.058 & 52.08\% \bigstrut[t]\\
    NeRF Fine-tuning & 27.07 & 0.924 & 0.073 & 83.98\% \bigstrut[t]\\
    \hline 
    \end{tabular}
    }
    \label{tab:rebuttal}
    
\end{table}

\subsection{Robustness}
Robustness is a critical metric for digital watermarking, aiming to ensure the signature can withstand various modifications or attacks without destruction or removal. We evaluate the robustness of our NeRF Signature and the baselines by subjecting them to different types of attacks.

% Table generated by Excel2LaTeX from sheet 'Sheet1'
\begin{table*}[htbp]
  \centering
  \caption{Bit accuracy under various image-level transformations compared with the baselines. Results are presented for $16$ bits and are averaged across all instances within the Blender dataset. The best-performing methods are highlighted in \textbf{bold}.}
  \resizebox{\textwidth}{!}{
    \begin{tabular}{c|ccccccccc}
    \hline
    Method & No distortion & Rotation &  Cropping & Scaling & JPEG & Blurring & Noise & Brightness & Combined \bigstrut[t]\\
    \hline
    NeRF Signature & \textbf{99.98\%} & \textbf{99.90\%} & \textbf{98.62\%} & \textbf{99.94\%} & \textbf{99.78\%} & \textbf{99.86\%} & \textbf{99.87\%} & \textbf{99.89\%} & \textbf{97.11\%} \bigstrut[t]\\
    NeRF Signature (w/o DiL) & \textbf{99.98\%} & 49.41\% & 88.54\% & 90.66\% & 90.43\% & 87.28\% & 74.66\% & 82.34\% & 55.32\% \\
    CopyRNeRF~\cite{luo2023copyrnerf} & 90.32\% & 88.53\% & 88.07\% & 87.16\% & 81.66\% & 89.10\% & 88.36\% & 86.57\% & 82.45\% \\
    NeRF Fine-tuning & 94.09\% & 89.33\% & 87.82\% & 88.93\% & 86.00\% & 89.26\% & 87.02\% & 87.65\% & 80.74\% \\
    \hline
    \end{tabular}%
    }
  \label{tab:image_robust}%
\end{table*}%

\subsubsection{Image-level transformations} We first consider transformations on the rendered images, similar to those considered by previous 2D watermarking schemes~\cite{zhu2018hidden}. Specifically, as illustrated in~\Tref{tab:image_robust}, we examine various types of 2D transformations, including rotation, cropping, scaling, JPEG compression, blurring, noise, and brightness. The results indicate that our method is quite robust to different 2D distortions. Although CopyRNeRF~\cite{luo2023copyrnerf} achieves enhanced robustness against image transformations by integrating a distortion layer during training, its bit accuracy values are still lower than those of our NeRF Signature.

\begin{table}[tbp]
  \centering
  
  \caption{Bit accuracy under fine-tuning attacks in different settings. Results are presented for $16$ bits and are averaged across all instances within the dataset.}
  \resizebox{\linewidth}{!}
    {
  \begin{tabular}{c|c|cccc}  
    \hline
    \multicolumn{2}{c|}{Attack setting} & 0 epochs & 100 epochs & 300 epochs & 500 epochs \bigstrut\\
    \hline
    \multirow{2}[2]{*}{w/o CI} & w/ PK   & 99.98\% & 99.98\% & 99.97\% & 99.96\% \bigstrut[t]\\
          & w/o PK  & 99.98\% & 99.98\% & 99.98\% & 99.97\% \\
    \hline
    \multirow{2}[2]{*}{w/ CI} & w/ PK   & 99.98\% & 66.75\% & 56.44\% & 52.13\% \bigstrut[t]\\
          & w/o PK  & 99.98\% & 98.91\% & 98.06\% & 97.50\% \\
    \hline
    \end{tabular}%
    }
  \label{tab:finetune}%
    
\end{table}%

\subsubsection{Fine-tuning attacks}

Beyond transformations at the image level, attackers may target the NeRF model itself. Model fine-tuning represents a common attack strategy. We examine two attack scenarios: without clean images (w/o CI) and with clean images (w/ CI). In the w/o CI scenario, attackers use synthetically rendered images with random noise for fine-tuning the watermarked NeRF. In the w/ CI scenario, they use original, unmodified images. Each scenario is further distinguished by whether the attacker has access to the pose key (w/ PK) or not (w/o PK). Without the pose key, attackers can only fine-tune from random viewpoints. With the pose key, they can fine-tune using the specific viewpoints corresponding to the key.

The results for fine-tuning attacks are shown in~\Tref{tab:finetune}. The results indicate that when the attacker does not have access to clean images, regardless of whether they know the pose key, their fine-tuning attacks cannot reduce the bit accuracy of our NeRF Signature. The bit accuracy of NeRF Signature decreases after a certain number of attack rounds if the attacker knows the clean images corresponding to the actual pose key, but the conditions for such an attack are very stringent in practice. Our method maintains a high bit accuracy rate under fine-tuning attacks when the attacker does not know the pose key. This indicates that using the secret key can effectively enhance the robustness of digital watermarking against such fine-tuning attacks.

\begin{table}[tbp]
\caption{Results of adversarial attacks in different settings. The adversarial attacks are implemented by PGD-40. Results are presented for $16$ bits and are averaged across all instances within the dataset.}
\label{tab:my-table}
\resizebox{\columnwidth}{!}{%
\begin{tabular}{cc|cc|cc}
\hline
\multicolumn{2}{c|}{\multirow{2}{*}{Attack setting}}                                                               & \multicolumn{2}{c|}{$\delta_{\text{adv}}=0.1$} & \multicolumn{2}{c}{$\delta_{\text{adv}}=1.0$} \bigstrut\\ \cline{3-6} 
\multicolumn{2}{c|}{}                                                                                              & Bit Acc.        & PSNR        & Bit Acc.        & PSNR       \bigstrut[t]\\ \hline
\multicolumn{2}{c|}{No attack}                                                                                     & 99.98\%         & 43.31       & 99.98\%         & 43.31      \bigstrut[t]\\ \hline
\multicolumn{1}{c|}{\multirow{5}{*}{\begin{tabular}[c]{@{}c@{}}Attack with\\ random keys\end{tabular}}} & seed \#0 & 99.91\%          & 38.27       & 99.89\%         & 27.23    \bigstrut[t]\\
\multicolumn{1}{c|}{}                                                                                   & seed \#1 & 99.03\%          & 38.21       & 97.28\%                & 28.24      \\
\multicolumn{1}{c|}{}                                                                                   & seed \#2 & 98.75\%         & 37.50      & 92.28\%       & 30.59           \\
\multicolumn{1}{c|}{}                                                                                   & seed \#3 & 99.81\%         & 37.36       & 93.25\%      & 25.96           \\
\multicolumn{1}{c|}{}                                                                                   & seed \#4 & 99.56\%         & 37.60        & 94.00\%         & 31.45       \\ \hline
\multicolumn{2}{c|}{Attack with the actual key}                                                                    & 61.75\%         & 37.37            & 50.66\%                &  30.65          \bigstrut[t]\\ \hline
\end{tabular}%
}\label{tab:adv}
\end{table}
\subsubsection{Adversarial attacks}
The adversarial attack is a technique that attempts to fool the downstream models by adding imperceptible perturbations to the input, which can lead to the failure of the downstream models. We assess the robustness of NeRF Signature against adversarial attacks. We consider a scenario where the malicious user can access the signature extractor. The malicious user leverages gradient information to introduce adversarial perturbations into the model parameters, aiming to mislead the signature extractor and prevent it from correctly extracting information from rendered images. For digital watermarking, the attacker's goal is to make the signature extractor obtain random binary signatures, thereby achieving an accuracy rate that approximates 50\%. If adversarial samples are generated solely by maximizing the signature loss of the extractor as in~\Eref{eq:loss_m}, it is equivalent to targeting the opposite of the actual binary signature, which is meaningless for digital watermarking. Therefore, in the setup of our experiment, the attacker randomly selects a binary signature as the target to make the signature extractor obtain it. Formally, the process of conducting an adversarial attack on the watermarked NeRF can be concluded as follows:
\begin{align}
&\min_{\Delta \Theta}\quad \texttt{CE}\left(f_{\text{m}}(\mathcal{R}(\Theta+\Delta \Theta, \tilde{\mathcal{K}}), \theta_{\text{m}}), \mathbf{m}_{\text{random}}\right),\\
&\text { s.t. } \quad \left\|\Delta \Theta\right\|_p \leq \delta_{\text{adv}},
\label{eq:adv}
\end{align}
where $\mathcal{R}$ is the rendering operator as defined in~\Eref{eq:render_op}, $\texttt{CE}(\cdot,\cdot)$ represents the cross-entropy loss computed between the two inputs, $\tilde{\mathcal{K}}$ denotes the key utilized for the attack, $\Delta \Theta$ signifies the adversarial perturbation applied to the parameters of NeRF $\Theta$, $\left\|\cdot\right\|_p$ denotes the $\ell_{p}$-norm, and $\delta_{\text{adv}}$ is the maximum allowable perturbation. The random signature $\mathbf{m}_{\text{random}}$ is selected by the attacker as the target of the adversarial attack. We consider two scenarios. In one scenario, the attacker is aware of the actual key, in which case $\tilde{\mathcal{K}}=\mathcal{K}$. In the other scenario, the attacker does not know the real key and thus can only attack by a randomly guessed one as $\tilde{\mathcal{K}}=\mathcal{K}_{\text{guess}}$. In practice, we employ PGD-40~\cite{madry2018towards} to implement the adversarial attack described above.

\begin{figure}
  \centering
  \includegraphics[width=\linewidth]{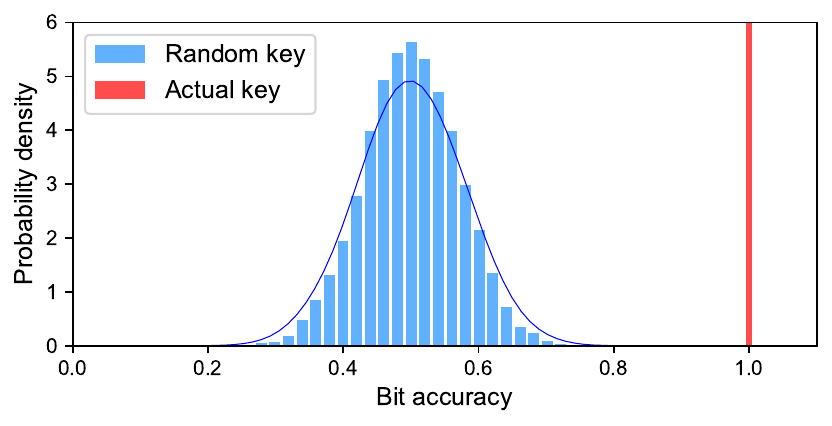}
  \caption{Results of attacks on the private key. We show a histogram of bit accuracies by using the same released extractor with either the correct key or $10000$ randomly chosen keys. Results are presented for $48$ bits on the ``hotdog'' case in the Blender dataset.}
  \label{fig:random_key}
\end{figure}

We show the results for adversarial attacks in~\Tref{tab:adv}. In our experiments, we choose two maximum allowable perturbations, epsilon, and select $5$ random seeds to generate the guessed keys. When the attacker can access the key, the results indicate that the attacked NeRF can effectively deceive the extractor into producing a random binary signature. However, when adversarial attacks are implemented using randomly guessed keys, the bit accuracy can remain very high. This proves that our proposed key mechanism can effectively resist malicious adversarial attacks. As maximum allowable perturbation $\delta_{\text{adv}}$ increases, the bit accuracies of random keys decrease slightly, but the rendering image quality, as measured by PSNR, decreases significantly. This suggests that our method is robust against this type of adversarial attack.

\subsection{Attacks on private key}

We evaluate the robustness of our method against private key attacks when malicious users have access to the extractor. The experiment simulates a scenario where attackers attempt to extract signatures without knowing the correct pose and patch keys. We conduct the experiment on a watermarked NeRF with a 48-bit signature.
    
As shown in~\Fref{fig:random_key}, we evaluate signature extraction using both the correct private key and $10,000$ randomly generated keys. With the correct key, our method achieves $100\%$ bit accuracy in signature extraction. However, attempts with random keys typically yield accuracies below $70\%$. This significant performance gap between correct and random keys demonstrates that our joint pose-patch encryption effectively prevents unauthorized signature extraction, even with extractor access.

\begin{table}
  \centering
  \caption{Training time compared with the baselines. Results are presented for $48$ bits and are averaged across all instances within the dataset.}
  \resizebox{\columnwidth}{!}{%
    \begin{tabular}{c|c|c|c|c}
    \hline
    \multicolumn{2}{c|}{Method} & NeRF Signature & CopyRNeRF~\cite{luo2023copyrnerf} & \begin{tabular}[c]{@{}c@{}}NeRF Fine-tuning \bigstrut[t]\\ (with fixed signature)\bigstrut[b]\end{tabular} \\
    \hline
    \multicolumn{2}{c|}{Training time} & $\sim 6$ minutes   & $>70$ hours  & $\sim 40$ minutes \bigstrut[t]\\
    \hline
    \end{tabular}%
    }
  \label{tab:time}%
\end{table}%

\begin{figure*}
  \centering
  \includegraphics[width=\linewidth]{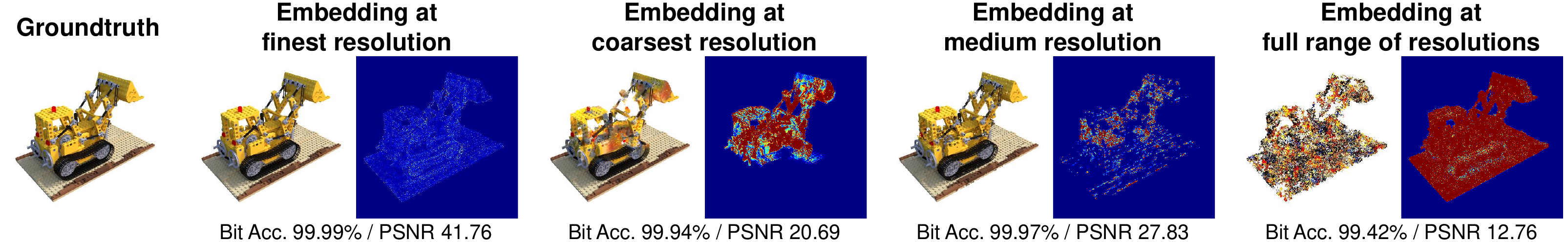}
  \caption{Ablation study on the choice of embedding portion of our method. Results are presented for $32$ bits on the ``lego'' case in the Blender dataset. The averaged Bit Acc. / PSNR is shown below each example}
  \label{fig:ablation_portion}
\end{figure*} 

\begin{figure*}
  \centering
  \includegraphics[width=\linewidth]{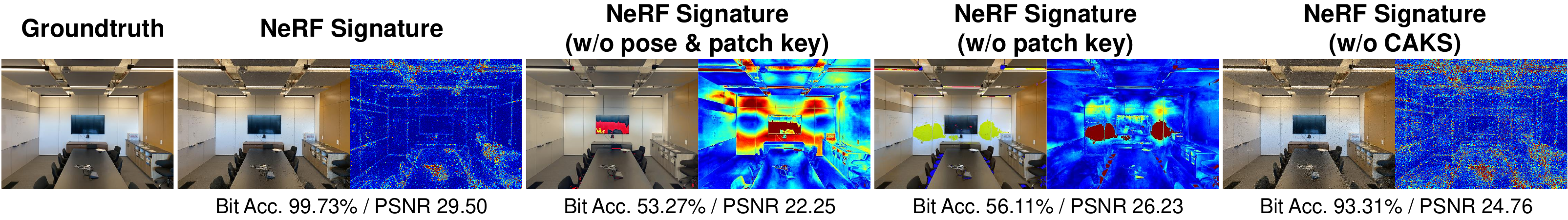}
  \caption{Ablation study on the design of our joint pose-patch encryption watermarking strategy and CAKS scheme. Results are presented for $48$ bits on the “room” case in the LLFF dataset. The averaged Bit Acc. / PSNR is shown below each example}
  \label{fig:ablation}
\end{figure*} 

\subsection{Training time}
We compare the training time required for our NeRF Signature with the training times for CopyRNeRF~\cite{luo2023copyrnerf} and NeRF Fine-tuning. The results are illustrated in~\Tref{tab:time}. We calculate the total time for all $2^{N_{\text{b}}}$ signature embedding situations. Since NeRF Fine-tuning can only embed one fixed signature at one time, we calculate the total time for $2^{N_{\text{b}}}$ fine-tuning sessions. The experimental results indicate that the NeRF Signature requires the shortest training time. Therefore, our method enables users to freely choose any signatures from $2^{N_{\text{b}}}$ possibilities to embed within a shorter training time.

\subsection{Computational efficiency}
The computational efficiency of the extractor is crucial for practical deployment, especially on edge devices. We evaluate the computational requirements of our extractor on a standard desktop environment with Intel Xeon CPU and NVIDIA Tesla V100 GPU.

For runtime performance, our extractor completes the signature extraction process in approximately $7.46$ ms. This efficiency is achieved through the use of lightweight convolutional operations in our architecture. Regarding resource consumption, the extractor has a small memory footprint with only $515$ KB runtime memory overhead. The storage requirement for the extractor model is $1023$ KB, making it suitable for deployment on resource-constrained devices.

The extractor's architecture primarily consists of standard convolutional operations that are well-supported by mobile deep learning frameworks. This design choice ensures compatibility with edge devices while maintaining extraction accuracy. Edge device deployment would facilitate practical applications by enabling on-device signature extraction.

\subsection{Ablation study}
Our approach consists of four parts: the signature codebook, the joint pose-patch encryption watermarking strategy, the complexity-aware key selection scheme, and the distortion layer. As the signature codebook is necessary for our method, we cannot easily remove it. However, we can assess the choice of embedding portion $\Theta_{\text{e}}$ as in~\Eref{eq:nerf_re}. Therefore, in our ablation study, we evaluate our method from the above-mentioned four aspects. 

\subsubsection{Effect of embedding portion choice}
In our main experiments, we choose the grid at the finest resolution in Instant-NGP~\cite{muller2022instant} for signature embedding. To evaluate the performance of other choices of the embedding portion, we conduct an ablation study by selecting different portions of the NeRF for signature embedding. The results are shown in~\Fref{fig:ablation_portion}. The results show that embedded signatures at the highest resolution can achieve the highest PSNR. Although embedded signatures at the lowest resolution, medium resolution, and full range of resolutions can all reach high bit accuracy, their visual quality drops a lot. The results indicate that choosing the grid at the finest resolution for signature embedding can maintain high imperceptibility.

\subsubsection{Effect of joint pose-patch encryption watermarking}
The core of our joint pose-patch encryption watermarking strategy is to use the pose and patch keys to embed and extract signatures. To demonstrate the effectiveness of this strategy, we test our method without applying the pose key and patch key. We first remove both the pose key and patch key, indicating that information is extracted across the entire rendered image from any viewpoint. Then, we use a pose key and remove the patch key to extract information from the whole rendered image from a specific viewpoint. The results are shown as NeRF Signature (w/o pose \& patch key) and NeRF Signature (w/o patch key) in~\Fref{fig:ablation}, respectively. These results indicate that using the pose key and patch key can effectively enhance the imperceptibility and effectiveness of watermarking.

\subsubsection{Effect of complexity-aware key selection scheme}
For patch selection, we apply a complexity-aware scheme to select patches with a higher complexity level of texture. To evaluate the effectiveness of such a key selection scheme, we conduct experiments by removing it. We randomly choose $N_{\text{b}}$ patches as the regions for embedding and extracting signatures. The results are shown as NeRF Signature (w/o CAKS) in~\Fref{fig:ablation}. The results demonstrate that utilizing a CAKS scheme enables a more imperceptible embedding of signatures in NeRF.

\subsubsection{Effect of distortion layer}
Our approach employs the distortion layer during optimization to enhance the system's robustness. To validate the effectiveness of the distortion layer, we compare the robustness against image-level transformation attacks between using and not using the distortion layer. The results of our method without the distortion layer are shown as Proposed (w/o DiL) in~\Tref{tab:image_robust}. The results demonstrate that the distortion layer is important in enhancing robustness against image-level degradation.

\section{Conclusion, Limitations, and Future Work}
We propose a novel watermarking method for NeRF called NeRF Signature, which offers high imperceptibility and robustness at both the image and model levels, while allowing the NeRF owner to flexibly choose the signatures to embed. First, using a codebook-aided signature embedding in our method can keep the NeRF structure unchanged during signature embedding without introducing extra complex modules. The NeRF owner can generate watermarked NeRF with any desired signatures through the signature codebook. Second, the joint pose-patch encryption watermarking strategy hides signatures in a secret pose and several patches, preventing the leakage of the signatures even when the extractor is publicly known. This also increases the difficulty of their attacks on targeted pose and patches if malicious users are unaware of the secret key.  Third, the CAKS scheme is proposed to select patches with high visual complexity, in which hiding signatures can cause less visual difference. Experiments on standard datasets show that our method outperforms other baselines in terms of bit accuracy and imperceptibility. Our method also demonstrates strong robustness against image-level transformations, model-level modifications, and attacks on the private key.

\noindent{\textbf{Limitations and future work.}} 
Our approach has demonstrated promising performance in asserting ownership of NeRF, a widely-used method for 3D representation that has served as the basis for numerous applications. However, the landscape of 3D representation methods is constantly evolving. For instance, 3D Gaussian Splatting (3DGS)~\cite{kerbl3Dgaussians} is an emerging 3D representation method that is rapidly developing. Due to the point cloud representation of 3DGS~\cite{kerbl3Dgaussians}, directly applying our approach would lead to a significant increase in storage space required for the signature codebook. Furthermore, the development of watermarking for 3D representations faces several key challenges. First, it is essential to develop scalable watermarking methods adaptable to different 3D representations like NeRF~\cite{nerf2020, muller2022instant} and emerging techniques such as 3DGS~\cite{kerbl3Dgaussians, yang2024gaussianobject, luiten2024dynamic}. Second, seamless integration into existing 3D content creation and distribution pipelines is crucial, which includes minimizing computational overhead and optimizing for edge devices while maintaining compatibility with downstream processes. Third, enhancing security and robustness against various attacks remains critical for protecting intellectual property. Addressing these challenges will foster a more secure and innovative ecosystem for 3D digital content creation and distribution.

\section*{Acknowledgments}
Renjie Group is supported by the National Natural Science Foundation of China under Grant No. 62302415, Guangdong Basic and Applied Basic Research Foundation under Grant No. 2022A1515110692, 2024A1515012822, and the Blue Sky Research Fund of HKBU under Grant No. BSRF/21-22/16. This research is also supported by Horus project number \#2023/12865-8, CNPq \#302458/2022-0 and CNPq Aletheia \#442229/2024-0. This research is also supported by the NSFC under Grant No. 62136001. This research is also supported by the National Research Foundation, Singapore and Infocomm Media Development Authority under its Trust Tech Funding Initiative. Any opinions, findings and conclusions or recommendations expressed in this material are those of the authors and do not reflect the views of the National Research Foundation, Singapore, and Infocomm Media Development Authority. This research is also supported in part by Sichuan Science and Technology Program under Grant No. 2025ZNSFSC0511.

{
\bibliographystyle{IEEEtran}
\bibliography{ref}
}

\section{Biography Section}

\begin{IEEEbiography}
[{\includegraphics[width=1in,height=1.25in,clip,keepaspectratio]{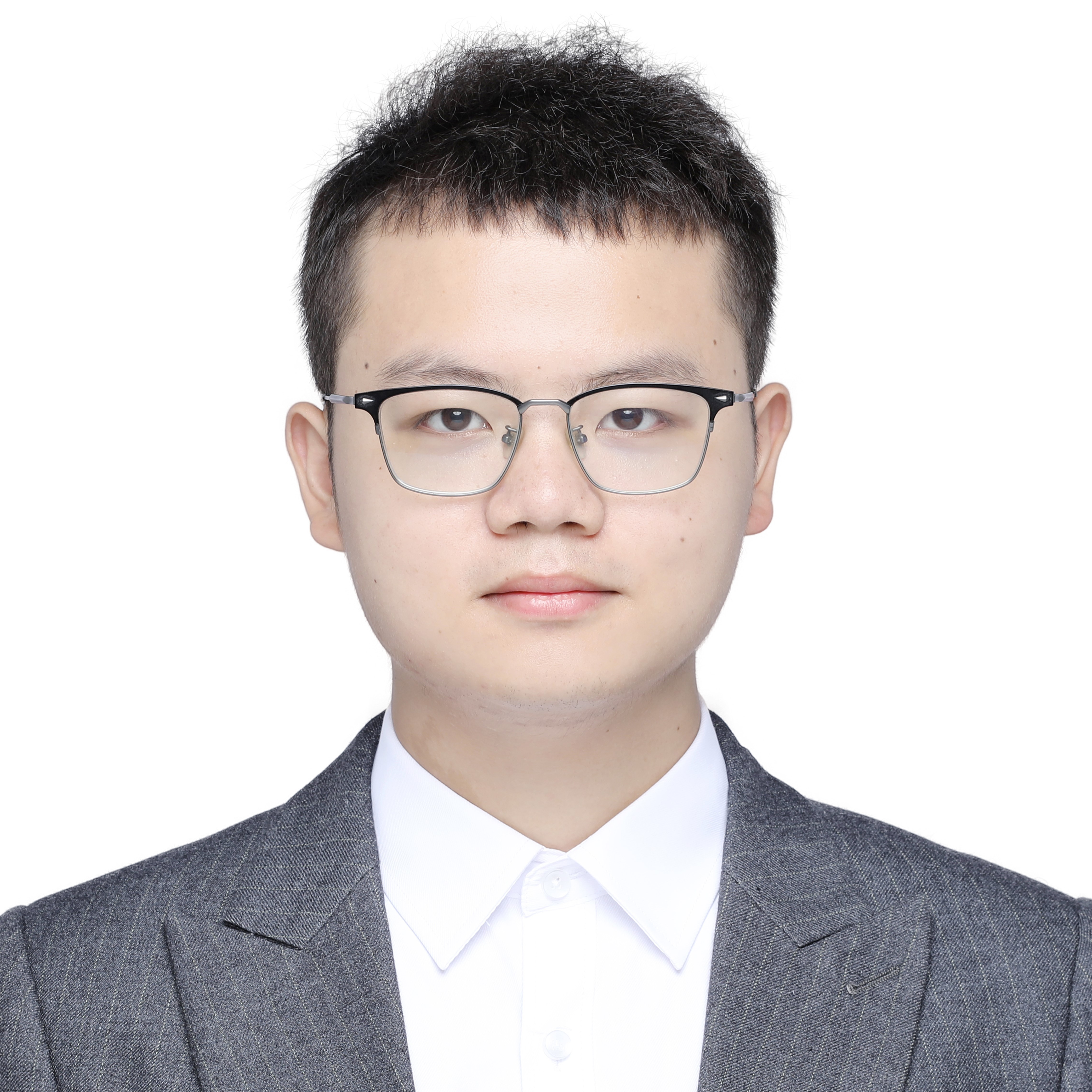}}]{Ziyuan Luo} (Student Member, IEEE)
is currently pursuing a Ph.D. degree at the Department of Computer Science, Hong Kong Baptist University (HKBU). He received his bachelor's degree and master's degree in Communication Engineering from the University of Electronic Science and Technology of China (UESTC), in 2019 and 2022, respectively. His current research interests include digital watermarking, 3D reconstruction, and AI security.
\end{IEEEbiography}

% \vspace{10pt}

\begin{IEEEbiography}[{\includegraphics[width=1in,height=1.25in,clip,keepaspectratio]{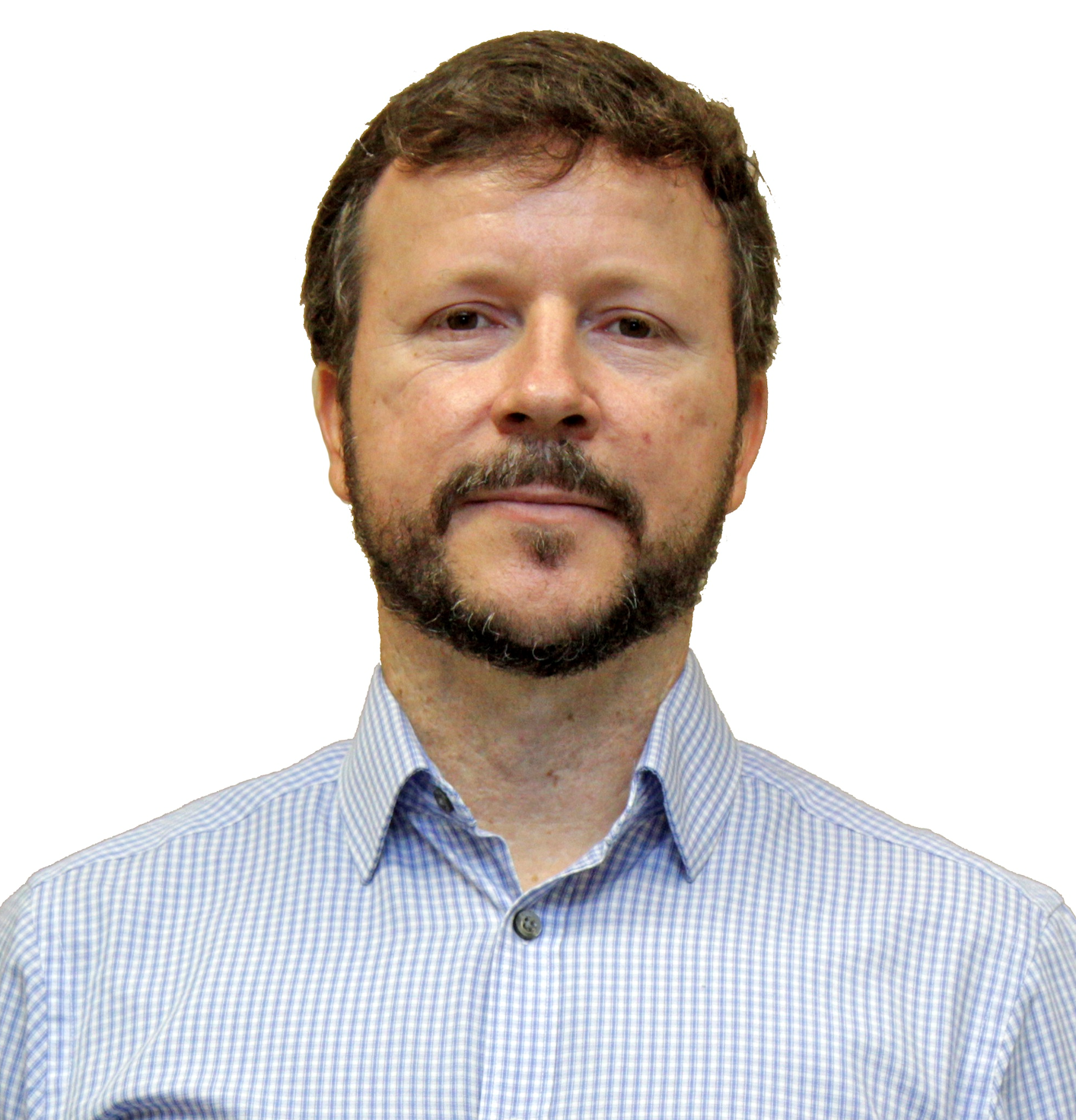}}]{Anderson Rocha} (IEEE Fellow) is Full-Professor of Artificial Intelligence and Digital Forensics at the Institute of Computing, University of Campinas (Unicamp), Brazil. His main interests include Artificial Intelligence, Digital Forensics, Reasoning for Complex Data, and Machine Intelligence. He is the Head of the Artificial Intelligence Lab., Recod.ai, at Unicamp and was the former Director of the Institute for the 2019-2023 term. He is an elected affiliate of the Brazilian Academy of Sciences (ABC) and the Brazilian Academy of Forensic Sciences (ABC). He is a three-term elected member of the IEEE Information Forensics and Security Technical Committee (IFS-TC) and a former chair of such committee. In 2023, he was elected again to the IFS-TC chair for the 2025-2026 term. He has actively been an editor of important international journals and the chair of key conferences in Al and digital forensics. He is a Microsoft Research and a Google Research Faculty Fellow, important academic recognitions given to researchers by Microsoft Research and Google. In addition, in 2016, he was awarded the Tan Chin Tuan (TCT) Fellowship, a recognition promoted by the Tan Chin Tuan Foundation in Singapore. Since 2023, he is also an Asia Pacific Artificial Intelligence Association Fellow. He has been the principal investigator of several research projects in partnership with public funding agencies in Brazil and abroad and national and multi-national companies, having already filed and licensed several patents. He is a Brazilian CNPq research scholar (PQ1C). He is ranked among the Top 2\% of research scientists worldwide, according to PlosOne/Stanford and Research.com studies. Finally, he is now a LinkedIn Top Voice in Artificial Intelligence for continuously raising awareness of Al and its potential impacts on society at large.
\end{IEEEbiography}

\vspace{6pt}

\begin{IEEEbiography}[{\includegraphics[width=1in,height=1.25in,clip,keepaspectratio]{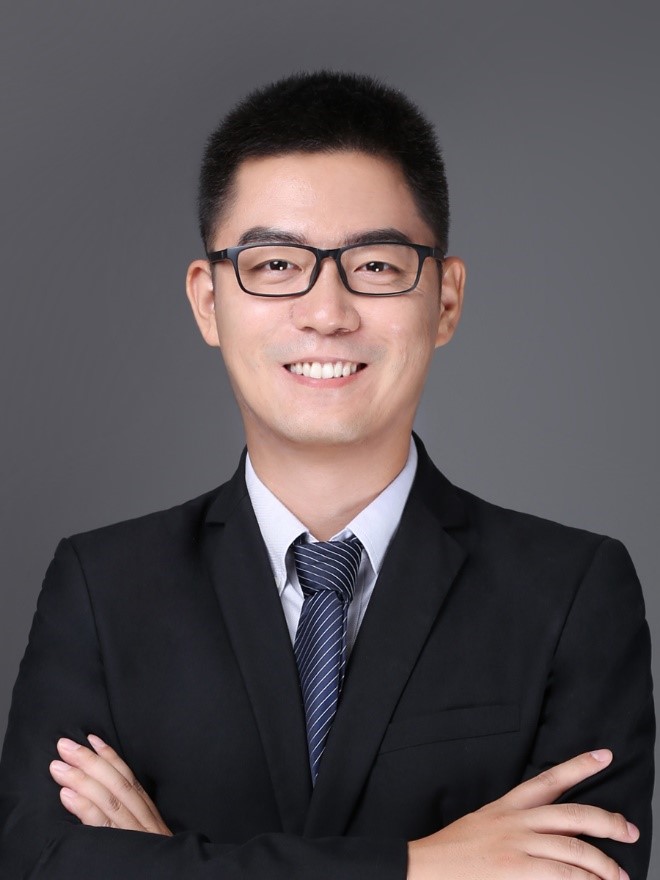}}]{Boxin Shi} (Senior Member, IEEE) received the BE degree from the Beijing University of Posts and Telecommunications, the ME degree from Peking University, and the PhD degree from the University of Tokyo, in 2007, 2010, and 2013. He is currently a Boya Young Fellow Associate Professor (with tenure) and Research Professor at Peking University, where he leads the Camera Intelligence Lab. Before joining PKU, he did research with MIT Media Lab, Singapore University of Technology and Design, Nanyang Technological University, National Institute of Advanced Industrial Science and Technology, from 2013 to 2017. His papers were awarded as Best Paper, Runners-Up at CVPR 2024, ICCP 2015 and selected as Best Paper candidate at ICCV 2015. He is an associate editor of TPAMI/IJCV and an area chair of CVPR/ICCV/ECCV. He is a senior member of IEEE.
\end{IEEEbiography}

\vspace{6pt}

\begin{IEEEbiography}
[{\includegraphics[width=1in,height=1.35in,clip,keepaspectratio]{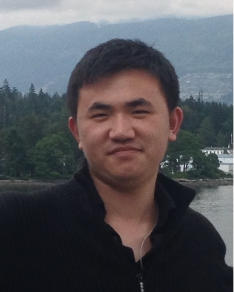}}]{Qing Guo} (Senior Member, IEEE) received a Ph.D. degree in computer application technology from the School of Computer Science and Technology, Tianjin University, China. He was a research fellow with the Nanyang Technology University, Singapore, from Dec. 2019 to Aug. 2020 and Dec. 2021 to Sep. 2022. He was assigned as the Wallenberg-NTU Presidential Postdoctoral Fellow with the Nanyang Technological University, Singapore, from Sep. 2020 to Dec. 2021. He is currently a senior scientist and principal investigator at the Center for Frontier AI Research and the Institute of High Performance Computing (IHPC), Agency for Science, Technology, and Research (A*STAR), Singapore. He is also an adjunct assistant professor at the National University of Singapore (NUS). His research interests include computer vision, AI security, and image processing.
\end{IEEEbiography}

\vspace{6pt}

\begin{IEEEbiography}[{\includegraphics[width=1in,height=1.35in,clip,keepaspectratio]{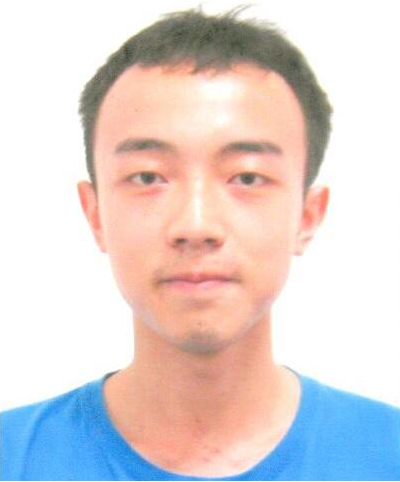}}]{Haoliang Li} (Member, IEEE)
received the B.S. degree in communication engineering from the University of Electronic Science and Technology of China (UESTC), Chengdu, China, in 2013, and the Ph.D. degree from Nanyang Technological University (NTU), Singapore, in 2018. He is currently an Assistant Professor with the Department of Electrical Engineering, City University of Hong Kong. His research interests include AI security, multimedia forensics, and transfer learning. He received the Wallenberg-NTU Presidential Postdoctoral Fellowship in 2019, the Doctoral Innovation Award in 2019,  the VCIP Best Paper Award in 2020, and ACM SIGSOFT distinguish paper award in 2022.
\end{IEEEbiography}
% \bf{If you will not include a photo:}\vspace{-33pt}
% \begin{IEEEbiographynophoto}{John Doe}
% Use $\backslash${\tt{begin\{IEEEbiographynophoto\}}} and the author name as the argument followed by the biography text.
% \end{IEEEbiographynophoto}
\vspace{6pt}

\begin{IEEEbiography}[{\includegraphics[width=1in,height=1.25in,clip,keepaspectratio]{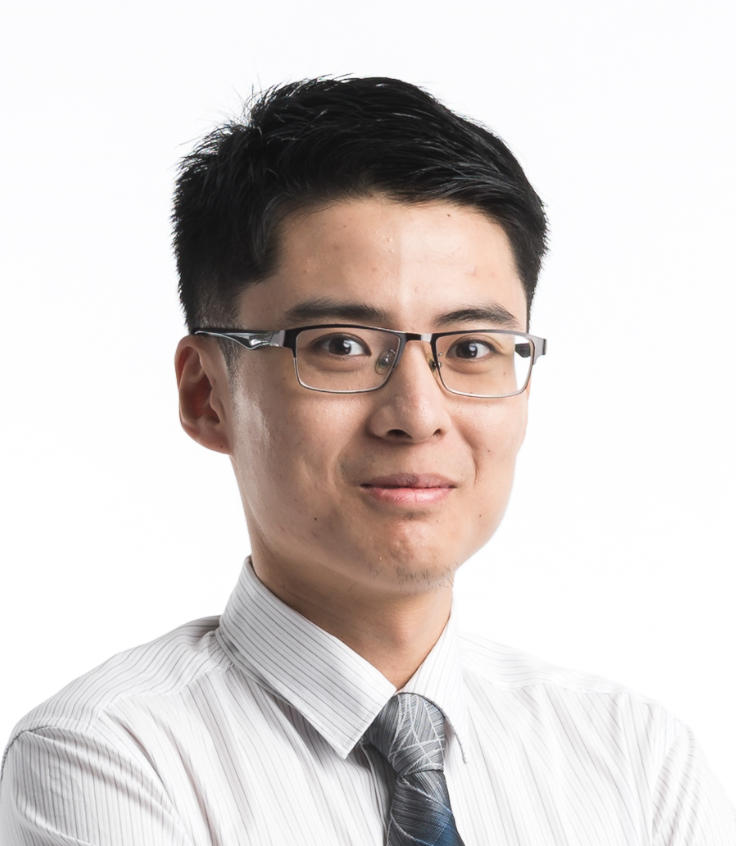}}]{Renjie Wan} (Member, IEEE) received the B.Eng. degree from the University of Electronic Science and Technology of China in 2012 and the Ph.D. degree from Nanyang Technological University, Singapore, in 2019. He is currently an Assistant Professor with the Department of Computer Science, Hong Kong Baptist University, Hong Kong. He was a recipient of the Microsoft CRSF Award, the 2020 VCIP Best Paper Award, and the Wallenberg-NTU Presidential Postdoctoral Fellowship. He is the outstanding reviewer of the 2019 ICCV.
\end{IEEEbiography}

\vfill

\end{document}